# H&E-based Computational Biomarker Enables Universal *EGFR* Screening for Lung Adenocarcinoma

Gabriele Campanella[1,2,3], David Ho[1], Ida Häggström[4,5], Anton S Becker[5], Jason Chang[1], Chad Vanderbilt[1], Thomas J Fuchs[1,2,3]

1. Department of Pathology, Memorial Sloan Kettering Cancer Center, New York, USA
2. Department of AI and Human Health, Icahn School of Medicine at Mount Sinai, New York, USA
3. Hasso Platner Institute for Digital Health at Mount Sinai, Icahn School of Medicine at Mount Sinai, New York, USA
4. Department of Electrical Engineering, Chalmers University of Technology, Gothenburg, Sweden
5. Department of Radiology, Memorial Sloan Kettering Cancer Center, New York, USA

## Abstract

Lung cancer is the leading cause of cancer death worldwide, with lung adenocarcinoma being the most prevalent form of lung cancer. *EGFR* positive lung adenocarcinomas have been shown to have high response rates to TKI therapy, underlying the essential nature of molecular testing for lung cancers. Despite current guidelines consider testing necessary, a large portion of patients are not routinely profiled, resulting in millions of people not receiving the optimal treatment for their lung cancer. Sequencing is the gold standard for molecular testing of *EGFR* mutations, but it can take several weeks for results to come back, which is not ideal in a time constrained scenario. The development of alternative screening tools capable of detecting *EGFR* mutations quickly and cheaply while preserving tissue for sequencing could help reduce the amount of sub-optimally treated patients. We propose a multi-modal approach which integrates pathology images and clinical variables to predict *EGFR* mutational status achieving an AUC of 84% on the largest clinical cohort to date. Such a computational model could be deployed at large at little additional cost. Its clinical application could reduce the number of patients who receive sub-optimal treatments by 53.1% in China, and up to 96.6% in the US.

## Introduction

Lung adenocarcinoma is the most prevalent form of lung cancer and has been found to have multiple somatic mutations that are amenable to tyrosine kinase inhibitor (TKI) therapy[1]. *EGFR* testing is considered mandatory by the CAP/AMP/IASLC guidelines[2] because of the high response rate of tumors to *EGFR* TKI. Current guidelines only allow for the performance of sequencing to identify *EGFR* mutations as a biomarker for TKI treatment.

Despite being considered necessary for the standard of care in lung cancer, molecular profiling of a tumor is not routinely performed on 24-28%[3,4] of lung cancer cases in the United States. The reason for the discrepancy between clearly published guidelines and actual clinical practice is not well understood but likely related to technical hurdles in obtaining and processing samples for testing as well as wide variations in clinical practice in the United States. Additionally, genomic sequencing, including targeted *EGFR* assays, is even less common in most regions of the world[5]. Given the high prevalence of *EGFR* mutation in lung adenocarcinoma, the lack of *EGFR* testing results in millions of patients receiving suboptimal therapy for *EGFR* mutated lung adenocarcinoma every year.

Even in properly resourced centers that have adopted standards for universal *EGFR* testing for lung cancer, many samples fail to be assessed for *EGFR* status due to insufficient material being available from diagnostic biopsies. Lung biopsies are inherently minute given the challenge of safely acquiring lung tissue. Additionally, the number of protocols necessary for a proper diagnosis and to obtain comprehensive tissue biomarker testing is large and ever expanding: PDL1 IHC, TTF-1 IHC, p40 IHC, ALK IHC, rapid *EGFR* testing, and broad genomic sequencing. Turnaround times (TAT) are a significant challenge for treatment of lung adenocarcinoma, especially for broad genomic sequencing with next generation sequencing (NGS) which has a TAT of approximately 2 weeks from the date of the biopsy. First-line therapy is typically withheld while waiting to receive *EGFR* mutation status as: (i) *EGFR* mutant tumors benefit from first-line TKI rather than chemotherapy/immunotherapy, and (ii) *EGFR* results further inform the likelihood of response to immune-based therapies. Rapid TAT testing has been developed and implemented to overcome this fundamental limitation of NGS. Unfortunately, these rapid TAT tests have important limitations as targeted tests they fail to detect less common *EGFR* variants, including *EGFR* exon 20 insertions that currently have Food and Drug Administration (FDA) approved treatments[6]. Most problematic for the rapid TAT tests is the need for additional sections of tissue or extracted DNA which results in an increased number of DNA quantity not sufficient (DQNS) failures for NGS tests and additional biopsies are necessary to obtain the remaining biomarker testing results.

There is a clear need for orthogonal methods to detect *EGFR* mutations that can be deployed with little cost, rapid TAT and simple implementation, while preserving tissue for broad genomic sequencing. Such a method would improve detection of *EGFR* mutant lung cancer across populations and dramatically reduce the number of sub-optimal treatment regimes. Detecting mutations directly from H&E slides, if adequate performance characteristics are achieved, offers a unique opportunity to overcome many of the limitations of current clinical sequencing for *EGFR*. A computational *EGFR* mutation detector would use as substrate only the digitized pathology slides from the diagnostic H&E biopsy with little overhead in terms of cost and physical processing for clinicians. Furthermore, such technology can produce results in seconds, which is attractive in situations where time constraints are important. Most importantly, a computational tool could be deployed anywhere in the world, making *EGFR* testing more accessible to millions of people in underserved regions of the world.

Prior studies have shown that molecular biomarkers for somatic mutations can be predicted directly from routine H&E slides[7,8]. Some recent examples include SPOP in prostate cancer[9], various mutations in lung adenocarcinoma[10,11], HPV driven head and neck cancers[12], and EBV driven gastric cancer[12]. Subsequent works have expanded the scope and searched for predictable mutations across cancer types and a plethora of genetic alterations[13,14]. In the context of *EGFR* prediction in lung adenocarcinoma, the seminal work from Coudray et al.[10] reported early results on the value of histology for this task. Beyond pathology, a growing number of studies are pointing to positron emission tomography (PET) imaging with 18F-fluorodeoxyglucose (FDG) bearing some predictive power for *EGFR* mutation status[15–19]. Additionally, it is known that some clinical variables (e.g., smoking status) are good predictors of *EGFR* status.

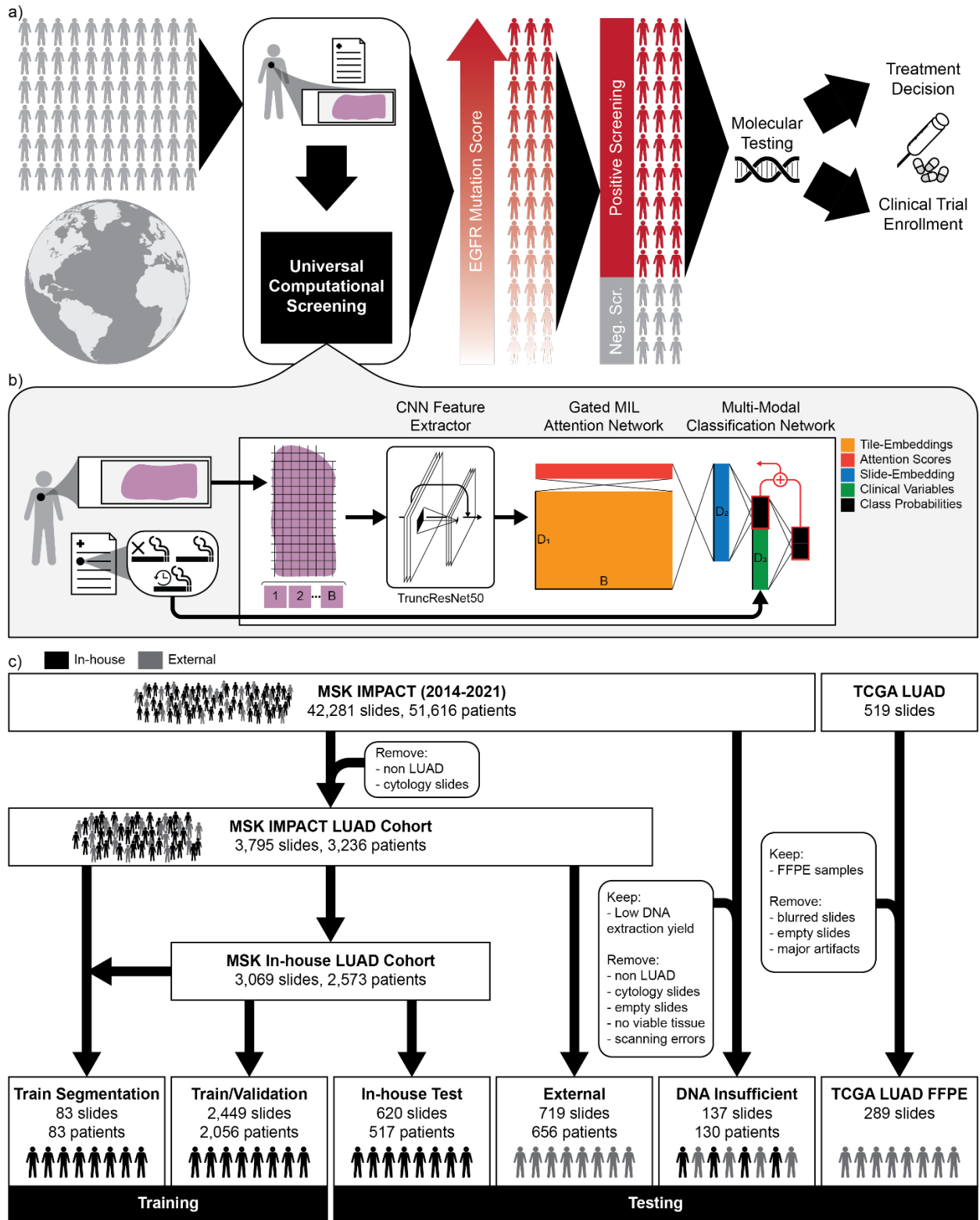

*Figure 1 Universal computational screening for EGFR positive lung cancers. a) The proposed Computational Biomarker can be deployed for global screening, given a digitized H&E slide and clinical notes as input. The EGFR score from the model can be used to prioritize molecular testing for optimal treatment decisions and accelerated clinical trial enrollment. b) Schematic of the proposed system. The model takes a pathology image and smoking status as inputs and outputs the probability of the tumor having and EGFR mutation. The model consists of a pretrained CNN for feature extraction, and a gated MIL attention-based*

*classification network which integrates pathology and clinical variables. $B$ is the number of tiles in a particular slide, $D_1 = 1024$, $D_2 = 512$, $D_3 = 1$. c) Patient cohorts analyzed in this study. Two main data sources were used, MSK and TCGA slides. The MSK cohort contained both in-house and external samples. The MSK in-house cohort was used for training and testing, meanwhile all other cohorts were used for testing only.*

In this work we expand on the evidence presented earlier that *EGFR* mutational status can be predicted with reasonable accuracy from morphological information in digitized pathology slides. We investigate the source of the *EGFR* signal and hypothesize a relation with morphological subtypes. We conduct extensive experiments in support of our hypothesis and investigate the integration of clinical variables and other modalities in the prediction of *EGFR* which can further increase performance. We also conduct an in-depth analysis of the "The Cancer Genome Atlas" (TCGA) lung adenocarcinoma (LUAD) dataset[20] to demonstrate that results obtained from it should be taken with caution. Finally, we discuss the benefits and impact of the clinical implementation of a computational biomarker for the universal screening of *EGFR* mutational status **(Figure 1a)**.

# Methods

## MSK–IMPACT Sequencing

To develop a computational biomarker for *EGFR* mutations, we obtained a large-scale cohort of lung adenocarcinoma patients at Memorial Sloan Kettering Cancer Center with respective molecular analysis from the MSK-IMPACT assay[21,22] and corresponding digitized molecular slides (see below). Briefly, MSK-IMPACT is the institutional hybridization capture based NGS assay performed routinely to detect clinically relevant somatic mutations, copy number alterations, and fusions in cancer. The assay detects variants in up to 505 unique cancer genes, including *EGFR*, in all tumor types. The analysis was restricted to formalin-fixed paraffin-embedded tissue samples (e.g., excluding cytology) and categorized as lung adenocarcinoma by a board-certified pathologist. All samples are sequenced in a CLIA certified laboratory, and all variants are reviewed by a board-certified molecular pathologist. *EGFR* mutations are clinically characterized using the OncoKB database[23]. Importantly, mutations outside of the *EGFR* kinase domain (exons 18-24) are not considered oncogenic and are excluded from the analysis. Oncogenic *EGFR* mutations are grouped into the common subtypes: (i) exon 19 deletions, (ii) L858R, (iii) exon 20 insertions, (iv) T790M and (v) other kinase domain mutations.

## WSI Datasets

To produce the image data, the last section taken from each formalin-fixed paraffin-embedded tissue block (after all unstained slides used for molecular sequencing have been cut) was stained with H&E and digitized with an Aperio AT2 digital slide scanner from Leica Biosystems at 20x magnification. The resulting cohort of 3,788 slides and 3,229 patients includes in-house and external cases (from 21 US states and 15 countries) with primary and metastatic samples (**Figure 1c, Extended Table 1**). The in-house cases were split into training and testing cohorts, whereas all external cases were reserved for testing. In addition, the "MSK insufficient DNA" cohort consisting of 134 slides from 127 patients was collected to represent cases that have a low amount of tumor tissue which would preclude the sample from undergoing sequencing. We also included a publicly available cohort of lung cancer slides from TCGA[20]. The 519 TCGA lung adenocarcinoma (LUAD) formalin-fixed and paraffin-embedded (FFPE) slides were curated to remove cases with artifacts that completely obscured the morphology of the sample (n=289). An in-depth description of the curation procedure is included in the **Extended Methods** section. The curated TCGA cohort was used as an additional external test set to further assess the generalizability of the proposed model. To assess the usability of TCGA data alone for developing a computational biomarker, additional cross-validation experiments were performed and reported. A subset of 83 MSK slides was also used to train and validate a lung segmentation model. A more detailed description of the cohorts is given in the **Extended Methods**.

### Other Data Modalities

For the MSK cohorts (with exception of the "lung segmentation" and "DNA insufficient" cohorts), a set of clinical variables was extracted from the clinical notes: smoking status, race and ethnicity, sex, and morphology subtype. Data relating to overall survival, age at diagnosis, and stage were also collected for a subset of the cases. With the exception of smoking status, which has a known inverse correlation with *EGFR* mutations, other clinical variables were not highly predictive of *EGFR* status (**Extended Figure 1a-c**). For a subset of 724 patients that also had received a pre-treatment PET/CT scan within 4 months of treatment start, their PET images were obtained to train a PET-based predictor.

### Lung Segmentation

A lung segmentation model was trained using Deep Interactive Learning (DIaL)[24] with a Deep Multi-Magnification Network (DMMN)[25] that segments 13 types of tumor and non-tumor morphologies. The cancer morphologic subtypes are acinar, lepidic, micropapillary, solid and papillary. Regions where the tumor cells spread through air spaces (STAS) were also segmented. Non-tumor morphologies are histiocytes, lymphocytes, airspaces, fibrosis, necrosis, pen markings and background. This model was used to restrict analysis to tumor regions in a slide, but also to correlate performance with individual cancer morphologies and limit training to subtype specific regions.

### Model Training and Validation

The MSK train/validation cohort was randomly split at the patient level into training (80%) and validation (20%) sets 20 times, each split containing a distinct set of patients for training and validation. For each split, the complete training procedure was replicated three times to control for variation in stochastic gradient decent (SGD) model optimization. The replicate with the best validation AUC after training was kept. The distribution of AUC values from the 20 splits was used to estimate the generalization performance and compare performances across different experimental set-ups. Performance on the various held-out test sets was measured by averaging the predictions from the top 10 best models on the validation set (top-10 ensemble) and using bootstrapping to estimate confidence intervals.

Several distinct neural network architectures were evaluated, including convolutional neural networks (ResNet34, ResNet50[26], and truncated ResNet50[27]), for tile-supervised classification and tile-level feature extraction, gated multiple instance learning (MIL) attention (GMA)[28] and transMIL[29] networks for slide-level classification and fully connected layers for multimodal integration of clinical covariates. For further details, see the **Extended Methods**. Our final proposed model comprises a truncated ResNet50, a GMA network and a multimodal classification network (**Figure 1b**).

### Cancer Prevalence, *EGFR* Mutation Prevalence and *EGFR* Testing Statistics

To assess the impact of a real-world application of the proposed computational biomarker, we first analyzed the current state of cancer prevalence, *EGFR* mutation rates and *EGFR* testing rates for four representative countries across the world to gain a global perspective. Lung cancer prevalence was obtained via various national registries[30–33], and the fraction of adenocarcinoma obtained from IARC's cancer incidence report[34]. To retrieve *EGFR* mutation and molecular testing rates, we performed a thorough literature search[3–5,35–39], but data is scarce and varies widely. To deal with this uncertainty, we use a low-high range of these estimates. **Extended Table2** contains all gathered information and an in-depth explanation is provided in the **Extended Methods**.

# Results

## Development of a Computational Biomarker for *EGFR* Mutations

### Experimental Design

To develop an accurate *EGFR* predictor, we compared and evaluated several training strategies. The most widely used approach is tile-level supervision by assigning the slide-level target to each tile of a slide followed by slide-level aggregation via average pooling. This strategy can only work with highly curated datasets such as TCGA relying on high tumor purity samples. The issue can be mitigated by restricting the analysis to tumor regions only on each slide, which we tested by training a deep learning-based lung segmentation model[25]. The former two strategies assign the same importance to each tile analyzed. The third strategy tested is based on the gated MIL attention model[28] where the importance of each tile in a slide is learned and a slide-level prediction is produced without need of further aggregation. This last model does not consider the spatial arrangement of tiles in a slide. The fourth strategy we employed is a transformer-based model (transMIL[29]) which incorporates the relative position of each tile in as slide into the analysis. Finally, we explored the addition of other modalities, in particular PET imaging and numerous clinical variables.

### Performance of an H&E-based Computational Biomarker

As depicted in **Figure 2a**, the best validation performance was achieved by the gated MIL attention model (average AUC 0.79). The two strategies with tile-based supervision yielded inferior performance (average AUC 0.73 and 0.77 respectively), suggesting that certain regions of the slides are more relevant for prediction. Interestingly, the spatially aware model did not achieve better performance (average AUC 0.75) than the gated MIL attention model which discards spatial information. These results suggest that higher-order spatial morphology may not be as relevant as local morphology for *EGFR* prediction. A detailed list of all conducted experiments is presented in the **Extended Results**.

We investigated the performance of the H&E based GMA model on the MSK test sets resulting in an AUC of 0.78 for the in-house cohort and an AUC of 0.77 for the external test cohort (**Figure 2b**). These results are well in line with the validation results, demonstrating excellent generalization performance on both in-house and external cohorts. We also analyzed the model's performance in terms of various confounding factors. We observed that prediction on primary samples performed better than on metastatic samples (0.81 vs 0.72 AUC **Extended Figure 3c**), regardless of in-house or external status. Other noteworthy observations were in terms of smoking status where current smokers were the most easily predicted and never smokers were the least easily predicted. Additionally, slides with a prevalence of solid subtype cancer underperformed compared to other subtypes. Interestingly, when a TP53 mutation was present, the performance of the model was degraded. A complete analysis of confounding variables is provided in the **Extended Results**.

### Performance on the insufficient DNA dataset

Even at institutions with in-house comprehensive genomic testing, profiling can fail due to insufficient DNA yield in the molecular sample. At Memorial Sloan Kettering Cancer Center, approximately 10% of genomic sequencing requests fail for this reason. When no additional tumor tissue is available, the patient must be re-biopsied to obtain the *EGFR* status, which leads to additional delays in management decisions. A computational approach could prove useful in these scenarios where the amount of tissue would not allow traditional molecular approaches. To prove this point, we tested our model on a specially curated dataset of molecular cases for which the DNA yield was too low for a successful IMPACT panel test to be obtained. The *EGFR* status was obtained from samples where the same tumor was re-biopsied or genomic sequencing was performed on a subsequent resection specimen. The performance on this cohort was on par with that of the other test cohorts, with an AUC of 0.80 (**Figure 2b**). This result further highlights the robustness of the proposed computational approach, even for samples with very low tissue and tumor presence.

## Performance of the TCGA dataset

The TCGA dataset has been widely used in many computational pathology studies, and frequently for mutation prediction in several different tumor types[8,10,13,14]. It is common knowledge of pathologists familiar with the TCGA archive that the slides and the scanned images are of variable quality. For most publications, the cohorts are manually curated to eliminate the slides unsuitable for analysis. This slide curation can be subjective and limit the reproducibility of studies performed on TCGA slides. We performed a thorough analysis of the various artifacts present in the FFPE TCGA LUAD cohort and share our curation results in the **Extended Methods**. We measured the performance of the model on the various data sub-selections obtained during the slide curation (**Extended Figure 5i**). By including all 519 TCGA slides, the performance of our model is quite poor, with an AUC of 0.61. By removing slides with artifacts so severe as to completely impede a pathological review, we obtained a curated set of 289 slides achieving an AUC of 0.70 (**Figure 2b**), a drop in performance of 0.06 compared to the external test cohort. Despite the removal of 52% of the cohort, the slides that were not removed were still of marginal quality when compared to the high-quality images used in clinical practice.

Despite the poor quality and heavy curation of the TCGA dataset, it has been the main resource for training predictive models for biomarkers in the computational pathology community. We performed several training runs from the TCGA data to assess the variability in results. To this end, we split the full dataset 30 times at random into training, validation and testing sets. Experiments for each split were performed twice to quantify within-split training variability. A complete description of these experiments is provided in the **Extended Results**. In short, the performance on the validation and test sets was extremely variable across different splits (AUC between 0.4 and 0.9) with the split replicates showing similar performance (**Extended Figure 5a-h**). These are indications that the TCGA cohort is not suitable for training robust predictive models.

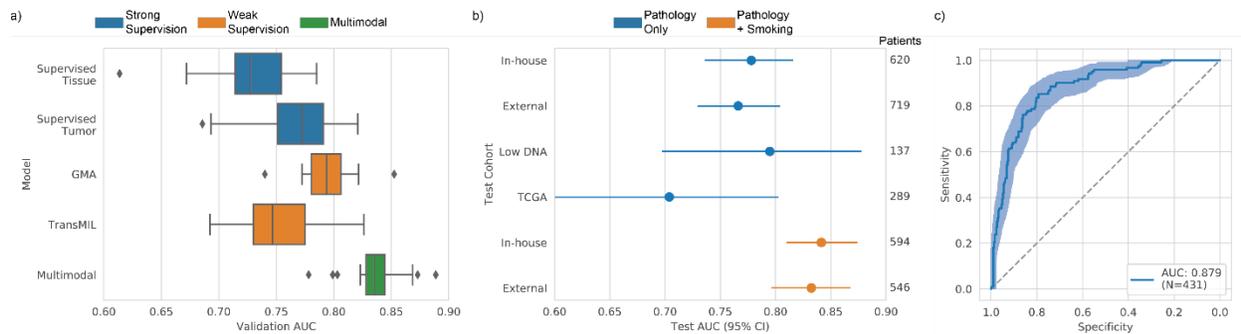

*Figure 2 Development and performance of a computational biomarker for EGFR mutation prediction. a) Performance comparison on the validation set: supervised training from all tissue, supervised training on tumor tiles only, Gated MIL Attention (GMA), TransMIL, and multimodal integration of smoking status with the histology GMA model. Each boxplot represents the performance variability from 20 random dataset splits. b) Performance comparison on all test sets for GMA models with and without smoking status: in-house, external, insufficient DNA, and TCGA. The Error bars represent the bootstrapped 95% confidence intervals. c) ROC curve of the proposed GMA + smoking status model for the in-house test cohort after removal of slides with less than 0.1 $cm^2$ of tissue. The shaded region represents the bootstrapped 95% confidence interval. This ROC curve is used for the subsequent clinical application analysis.*

## Importance of Cancer Subtype

To gain more insight into the model's representation of tissue morphology, we analyzed high attention regions in search of correlates with known biological concepts. We analyzed the attention maps generated by our trained top-10 ensemble model on the in-house test cohort. Importantly, we distinguished between attention for a positive and negative *EGFR* prediction (see **Extended Methods**). We correlated the median attention value within regions segmented computationally by morphological subtypes (acinar, papillary, micropapillary, lepidic, and solid). In **Extended Figure 3i** we observe a clear differential in positive attention for acinar, papillary, and micropapillary subtypes, for *EGFR* mutated slides versus *EGFR* wildtype slides, while solid and lepidic are much less represented in high attention regions.

To further investigate this point, we trained tile-supervision models (similar to the "all tissue" and "tumor tissue" supervised models described above) directly from specific subtypes only. The acinar, papillary, and micropapillary morphologies showed superior performance: 0.78, 0.75, and 0.76 average validation AUC respectively (**Extended Figure 2a**). The performance of each of these models was improved or on-par with the model that was trained on all tumor tiles and very similar to the performance obtained by the GMA model. This result is in line with other evidence discussed above that these specific subtypes contain most of the discriminative morphologies for prediction of *EGFR* mutation status. In contrast, models trained on solid and lepidic subtypes showed performance that was markedly degraded with 0.72 and 0.66 average validation AUC respectively. Thus, solid and lipidic subtype appear to have fewer discriminative morphologic traits.

## Integration of pathology, radiology and clinical modalities

Based on our previous experiments, we investigated the addition of various clinical variables to improve prediction performance. We experimented with the inclusion of various subsets of clinical variables and determined that including the smoking status enhances the validation performance with an average AUC of 0.84 (greater than using either pathology (0.79) or smoking status (0.70) alone). Further inclusion of up to 8 other clinical variables and even PET images does not improve performance further. See

**Extended Results** for a detailed recount of all experiments. The GMA-based pathology + smoking status is thus our final proposed model.

Due to the lack of smoking status on a subset of cases, in particular on external cases, we constrained the multimodal analysis to cases with known smoking status (in-house: 594 cases, external: 546 cases). For the in-house test cohort, the model achieved an AUC of 0.84, a performance well in line with the validation results, and on the external cohort the results were comparable with an AUC of 0.83 (**Extended Figure 4b**). Overall, the proposed model achieved an AUC of 0.84 on the combined test cohort of 1,140 samples. Stratification by primary and metastatic samples (**Extended Figure 4c**) demonstrated good generalization for primary cases from the in-house to the external cohort (AUCs of 0.85 and 0.86 respectively), mirroring the results of the histology-only model. The inclusion of smoking status improves the performance of the in-house metastatic samples (AUC of 0.84) to be on par with the primary samples, while the performance for the external metastatic samples remains lower (AUC of 0.80).

## Model Introspection and Error Analysis

We analyzed the histological traits used for prediction by projecting the embedding vectors into a two dimensional projection using UMAP[40]. One can observe that regions of the slide that the model has determined to have discriminatory characteristics (high attention) clustered together in the 2D projection. Furthermore, there was a clear topological distinction between clusters of regions with positive or negative attention. High attention regions of slides were visualized by a board-certified pathologist (C.V.). The tumor regions the model attended to for predicting *EGFR* alterations consistently have cells with large nuclei with open chromatin and prominent nucleoli. Regions that predict the tumor to be *EGFR* mutation negative have medium to large nuclei that are dark and consistently stained. This morphological difference between the positive and negative attention was observed independent of the morphologic subtype (**Figure 3**).

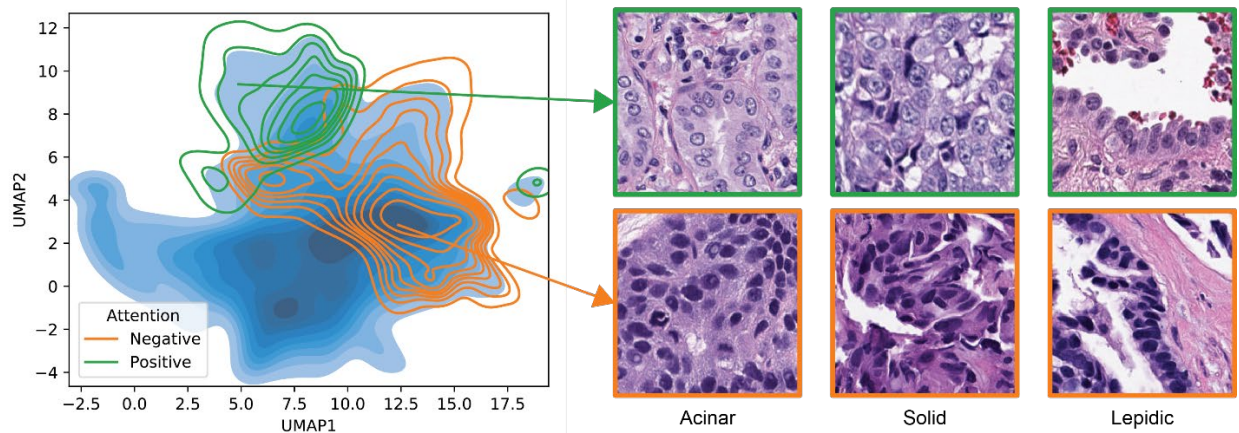

*Figure 3 Model introspection and UMAP analysis of learned histological features.* Left: 2D UMAP projection of feature vectors extracted from histology tiles by our trained CNN. The shaded surface represents the density of 459,768 tiles randomly sampled from 400 slides. Regions of the UMAP space with high positive and negative attention are shown with colored isolines. Right: Example tiles from regions of high positive and negative attention for acinar, solid, and lepidic subtypes. Positive attention tiles have cells with large nuclei with open chromatin and prominent nucleoli, whereas negative attention tiles have medium to large nuclei that are dark and consistently stained. This difference was observed independently of the morphologic subtype.

The whole slides that were predicted to be *EGFR* negative but were positive for mutations (false negatives) were analyzed in detail. The majority of these discrepant cases showed high cytologic grade, high mitotic count, and solid morphology. We observed that these false negative cases were enriched for TP53 mutations. Stratifying the test cohort by TP53 status we observed an improvement for the TP53 negative cohort with an AUC of 0.85 (compared to 0.83 for TP53 positive samples). This suggests that when *EGFR* mutant tumors are also TP53 mutated, some of the *EGFR* mutant discriminatory morphologic traits are lost. This finding also goes along with our previous finding that the model performs worse on lung adenocarcinoma with predominant solid pattern which are enriched in TP53 mutant tumors. As it is not possible to know a priori the TP53 mutation status the model's overall performance cannot be improved by incorporating the TP53 mutation status into the model.

## Universal Clinical Screening for *EGFR* Mutations

For the following analyses regarding the clinical application of the proposed model as a screening tool, we mimicked commercial products by adding a quality control step that filters out samples from the in-house test set containing less than 0.1 squared centimeters of tissue. This improved the performance of the model to an AUC of 0.88 (**Figure 2c**).

To determine the impact of a universal screening tool for *EGFR* mutations on patient treatment, we compare the number of sub-optimal lung adenocarcinoma treatments under the current standard of care to a hypothetical deployment of the screening tool. The number of current sub-optimal treatments is defined as the number of lung adenocarcinoma patients who have an *EGFR* mutation and do not receive molecular testing and thus no targeted therapy. Assuming universal adoption of the screening tool and compliance with a follow-up molecular test given its recommendation, we can calculate the number of positive screening results and the change in sub-optimal treatments for any combination of sensitivity and specificity (**Extended Figure 6**). The number of positive screening results increases with increased sensitivity as well as with decreased specificity of the model. The exact landscape depends on the number of LUAD cases and the prevalence of *EGFR* mutations in the patient population. The number of sub-optimal treatments only depends on the model sensitivity. At high sensitivity, the number of sub-optimal treatments is drastically reduced no matter the specificity level but would dramatically increase the number of subsequent molecular testing.

We model the scenario where the rate of molecular testing is kept constant at the current level of standard of care, but the molecular test is administered following the screening tool's recommendation. To do so, we determine the surface of the sensitivity/specificity plane that yields the predefined number of positive screening results (within 5% margin) and intersect it with the ROC curve from our model (**Figure 4a**). To consider the variability in the estimates for prevalence of *EGFR* mutations and molecular testing, we show best- and worst-case scenarios for each of the four countries included in the analysis.

To be more concrete, in the United States there are approximately 250,000 new lung cancer patients per year[30], 100,000 of which are adenocarcinomas. Given the current rate of molecular testing and the prevalence of *EGFR* mutations, one can estimate that between 2,214 and 6,601 patients in the US receive sub-optimal treatment due to the lack of molecular testing and targeted therapy (**Figure 4b-c**). Conversely, by keeping the numbers of performed tests fixed and following the model's prediction, the number of sub-optimally treated patients would decrease to a best case of 101 and a worst case of 644 (**Figure 4c**) without increase in molecular testing, representing a reduction of 90.7%--96.6% (**Figure 4d**). Similarly, for China we would see a decrease from 102,587--142,944 to 39,449--80,805 (53.1%--72.7% reduction). In Brazil we would observe a decrease from the current 569--1,992 to 135--843 (73.6%--84.1% reduction). Finally, in Germany the decrease would be from 838--1,066 to 94--154 patients (90.4% reduction). These numbers clearly demonstrate that a computational screening tool for directing molecular testing dramatically improves the current state of TKI treatment decision making.

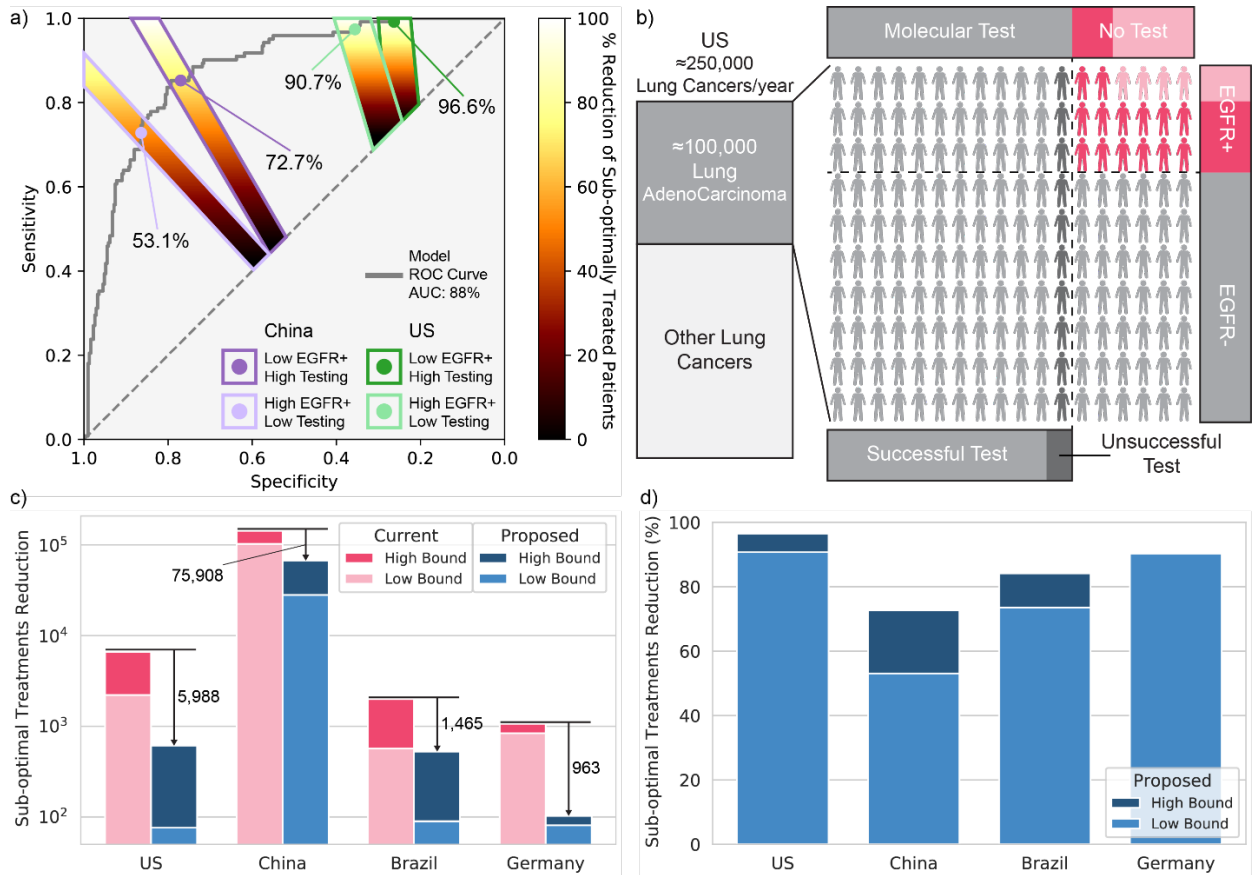

*Figure 4 Clinical impact of computational screening for EGFR mutations.* *Given the wide range of reported estimates of EGFR positive mutation prevalence and of molecular testing for EGFR in the literature, we report the lowest and highest estimates for each discussed scenario. a) Reduction of sub-optimally treated patients in the US and China when molecular tests are deployed based on the proposed computational screening tool (88% AUC), assuming the current number of molecular tests is kept fixed. Given a certain population of LUAD patients, the rate of EGFR mutations, and the rate of molecular testing, we identify the surface of the sensitivity/specificity space that corresponds to the number of molecular tests currently performed. Within each surface, the percentage reduction of non-optimal treatments is shown as a heatmap (c.f. Extended Figure 6). The intersection with the model's ROC curve shows the possible reduction achieved for each scenario. b) The current landscape in the US. Yearly incidence of lung cancer of 250k cases of which about 100k are adenocarcinomas. Given the current rates of molecular testing and the estimates of EGFR mutation prevalence, between 2.2% -- 6.4% of lung adenocarcinoma patients are not subjected to molecular testing although they have an EGFR mutation and hence receive a non-optimal treatment (patients shaded in pink). Furthermore, about 10% of molecular tests fail in practice due to insufficient DNA in the sample. c) Reduction in number of patients who receive non-optimal treatment using the proposed screening approach (blue) in contrast to the current standard of care (pink). d) Percent reduction of sub-optimally treated patients. The proposed screening prioritizes the molecular test based on our model's predictions. In both strategies, the same number of molecular tests are performed, based on current testing efforts for each country. The proposed screening strategy can significantly lower the number of patients who receive sub-optimal treatment.*

## Clinical Trial Enrollment Enrichment

The proposed model could also be used as a screening method for clinical trial enrollment. With the current strives in precision medicine it is often difficult to achieve sufficient patient enrollment for a clinical trial and the cost of sequencing can significantly impact investigators' ability to recruit patients. In the case of trials for the treatment of *EGFR* positive lung cancers, given the relatively low prevalence of these mutations, it is exceedingly difficult to obtain an adequate number of patients with *EGFR* mutant lung cancer. In the US, assuming an *EGFR* prevalence of 16% (average of low and high estimates[36]), random patient selection would discover on average only 16% of eligible patients from the candidate cohort. A computational screening based on the proposed model would result in a significant increase of *EGFR*-positive patients resulting in 50% enrichment without the need for additional molecular testing (**Figure 5a**). Given a predefined number of patients screened, the computational approach will yield higher *EGFR* positive patient rates. For example, if one clinical trial can only screen 1,000 patients due to budget constraints, a random screening method would return 142 eligible patients with 95% confidence, while the computational approach would uncover at least 443 patients with 95% confidence (**Figure 5b**). Given the prevalence of *EGFR* mutations in the US, the enrichment of eligible patients over a conventional, random screening approach would be roughly 3.2-fold. Brazil and Germany would have a similar improvement (3.1- and 3.4-fold respectively), while for China, given its higher prevalence of *EGFR* mutations, the enrichment of *EGFR* positive patients in screening would be 1.9-fold (**Figure 5c**). The proposed system can be deployed universally, making patient enrollment in clinical trials for *EGFR* positive cancers much more attainable.

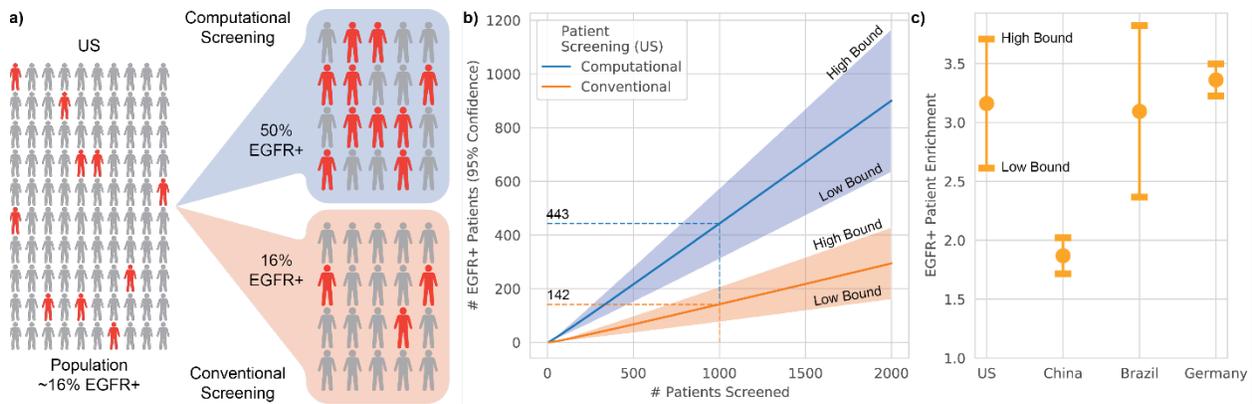

*Figure 5 Application of computational screening for patient enrollment in clinical trials.* *Given the wide range of estimates of EGFR positive mutation prevalence and of molecular testing for EGFR, we report the lowest and highest estimate in addition to the average for each scenario. a) Comparison of screening strategies for patient enrollment in clinical trials for the US. Random sampling of patients for molecular testing will result on average in the same percentage of EGFR positive patients as in the general population, while screening based on the computational approach will result in 3.2 times more suitable candidates on average. b) Given the number of patients enrolled in a trial (e.g., 1000) the plot shows the enrichment of EGFR positive patients for conventional selection (142 patients) versus computational screening with our model (443 patients). c) Fold-change in patient enrichment using computational screening compared to current random selection. Plotted is the average for each country including the lower and higher bound estimates.*

## Discussion

Currently, *EGFR* mutation testing is robust for patients that are able to receive molecular sequencing. The rates of false positives and false negatives are exceedingly rare given adequate DNA content and tumor proportion for NGS testing. The drawbacks are that NGS requires significant technical expertise, is costly to develop, and the TAT for the test, of about two weeks, can be inadequate for making optimal clinical decisions with regards to first line therapy. A few sophisticated laboratories have implemented stop gap testing where targeted PCR-based tests are used to give a rapid *EGFR* result while the remaining NGS panel is processed in the two-week timeframe. Given that the PCR-based tests require additional DNA, depleting the tissue and thereby decreasing the success rate of broad genomic sequencing, this strategy is clearly suboptimal. Furthermore, PCR tests do miss critical mutations that would inform clinicians of the tumor's responsiveness to TKI therapy. A method with rapid/instant TAT and sufficient accuracy would allow clinicians to start *EGFR* specific therapy early. The results of the computational model could then be confirmed by NGS. Most importantly the computational approach does not utilize any additional tissue and thus will not negatively impact NGS.

Despite the widespread and accepted understanding that testing for clinical actionable somatic mutations improves clinical care, there is a documented[3–5] and widespread lack of molecular testing across the globe. While estimates of molecular testing vary quite a lot, the inadequacy of the current efforts is apparent. In the US, a recent survey[3] determined the testing rate for *EGFR* alteration to be 72%, while another study found a rate of 76%[4]. While these estimates are reflective of regions that are served by elite medical centers, the situation looks much different for more under-served regions of the country. The disparity in access to molecular testing in the US is well reported[38], especially for next generation sequencing[39]. Unfortunately, in other countries the data can be even sparser. To gain a global view, we chose China, Brazil, and Germany, as reference for various regions of the world. Pennel et al.[5] report a testing rate for China between 42% and 46%, 38% for Brazil, and 66% for Germany. Overall, it is apparent that much is to be done to address the lack of testing which accounts for the thousands of patients who receive a sub-optimal treatment in the US every year, and hundreds of thousands per year in China alone.

A computational screening tool for targetable *EGFR* could be a turning point towards achieving optimal treatment strategies in lung cancer care. Previous studies have explored the notion of predicting certain biomarkers from histology. Notably, Coudray et al.[10] analyzed lung adenocarcinoma H&E slides and reported promising results for the prediction of *EGFR* in 2018. Unfortunately, these efforts have not bridged the gap from proof of concept to clinical application. We believe this is in part due to the reliance on TCGA datasets which are highly curated and do not translate well to real-world clinical scenarios. Further, some TCGA-based studies make full use of frozen sections which are not performed on routine diagnostic lung cancer biopsies. For the task of predicting *EGFR*, TCGA-based studies did not consider the functional implications of *EGFR* mutations outside the kinase domain which are not expected to respond to TKI therapy. In addition, most studies have relied on simple training strategies that assume the source of the signal is contained in small tiles and that all regions in a slide are equally important for prediction. Further research in how to extract relevant information from digitized slides is needed.

In this study we addressed all the previous points and presented a thorough analysis on the prediction of targetable *EGFR* mutations directly from H&E slides from the largest cohort of real-world clinical lung cancer specimens. This is only possible based on years of work and collection of genomic data from the sequencing efforts at a large referral cancer center. The scale of the work here reported, unlike any other of this kind, enables us to propose with confidence the first universal computational screening for *EGFR* mutations detection in lung adenocarcinoma utilizing H&E images.

Our analyses demonstrate that the application of the proposed *EGFR* mutation prediction model can direct a more efficient administration of molecular tests while dramatically improving treatment decision for lung adenocarcinoma patients. At the same time, its application will bolster patient inclusion in clinical trials, facilitating and accelerating the development of new treatments for *EGFR* positive lung adenocarcinomas. Since its use depends on digitized pathology slides and clinical variables such as smoking status, it can have a broad application at point of care. Computational methods like this have the possibility to reach patients where molecular testing so far has proven to be unfeasible. Additionally, it has little overhead costs since its input should be readily available, making it appealing in more economically constrained settings. Importantly, its turnaround time is lower than a minute and it does not exhaust tissue. We envision that widespread deployment of models such as this could reach universal adoption quickly because of their simplicity of use. The presented analyses demonstrate clearly that the impact on patients with lung cancer would be significant and in particular for patients in underserved communities without access to default molecular testing.

### Limitations

While the use of computational biomarkers for *EGFR* status prediction is very promising, there are some shortcomings that should be investigated further. In this work we focused on biopsies and excluded cytology samples, for which further work is needed to assess their suitability for computational *EGFR* prediction. The prediction performance could be improved further by integrating additional modalities in the model. The use of digital slides opens the possibility of deploying these systems much more widely than molecular testing, but currently there is still relatively low adoption of digital pathology worldwide. A bridge between the use of a computational biomarker and deployment in resource-constrained regions could be small desktop scanners or the digitization via microscope mounts for smartphone cameras or similar technologies. Additionally, the deployment to regions with challenges in slide preparation should be investigated, as certain types of slide artifacts may not have been observed during training.

## Conclusions

In this study we explored the development and application of a Computational Biomarker as a universal screening tool for targetable *EGFR* mutations in lung adenocarcinoma patients. The system is based on digitized pathology images and smoking status, is fast, can be easily deployed globally, and achieves reliable performance on an exceptionally large test dataset which includes tissue from 15 countries and 21 US states. A universal application of this screening tool would (i) direct more patients to a molecular test, in turn leading to optimized treatment, (ii) optimize the use of molecular testing in the population by prioritizing patients with high likelihood of being *EGFR* positive and (iii) facilitate the design of clinical trials for treatments of *EGFR* positive lung cancers. The increasing adoption of digital pathology is slowly but surely changing the landscape of medicine with the potential to improve care and outcomes for cancer patients and beyond. The recent FDA approval of an AI for diagnostic reporting in prostate cancer[41] hints at a not-so-distant future where AI supports clinicians to provide optimized care in a personalized manner. While computational pathology for diagnostics has reached a mature state, the next step in the field will bring computational tools for screening of molecular markers directly from H&E slides. We showed how a Computational Biomarker of this kind can be designed for the case of *EGFR* mutations in lung adenocarcinomas and how it would positively impact hundreds of thousands of patients globally.

# Extended Methods

## WSI Datasets

Development and testing of our deep learning models is based on a total of 4,508 slides from 3,875 patients divided into multiple cohorts as shown in **Figure 1c** and detailed in **Extended Table 1**: i) MSK in-house training cohort for training and validation; ii) MSK in-house test cohort and iii) MSK external test cohort for independent testing of model performance on biopsies taken at external institutions and submitted to MSK; iv) MSK insufficient DNA cohort for independent testing on slides with insufficient tumor content for molecular sequencing; v) MSK in-house segmentation cohort for training and validation of a lung cancer subtype segmentation model; vi) TCGA FFPE cohort for independent testing on a public dataset, and for exploring the feasibility of training on TCGA data.

## MSK In-house Cohort

For the in-house cohort, tissue was obtained from procedures performed at MSKCC. The H&E slides were also stained and scanned at MSKCC. A total of 2,573 patients and 3,069 slides were included. These comprised 2,201 primary samples and 868 metastatic samples. This cohort was split into a train set (2,056 patients, 2,449 slides) and a held-out test set (517 patients, 620 slides). To estimate the variability of the prediction performance, we randomly drew 80% of samples from the training set for learning a model and used the remaining 20% for validation. This procedure was repeated 20 times and the final distribution was reported as boxplots. For the weakly supervised experiments, the training set was randomly split in half, using one for feature learning and the other half for training the attention model.

## MSK External Cohort

We compiled a cohort of external cases to assess the generalizability of the method on non-MSKCC tissue. A slide is considered to be "external" if the slide comes from tissue that was procured from a procedure outside of MSKCC. These cases are from patients with lung adenocarcinoma who came to MSKCC for a second opinion or treatment after being managed at an outside hospital. The tissue was processed at the external facility and the FFPE tissue block was sent to MSKCC for sequencing. The unstained sections were used for DNA sequencing and the final slide was H&E stained and scanned at MSKCC. This cohort consists of 656 patients and 719 slides (460 primary samples and 259 metastatic samples). These patients come from 21 US states and 15 countries.

## MSK Insufficient DNA Cohort

The MSK-IMPACT assay requires a minimum of 25 ng of DNA to successfully generate results. Once a slide has been reviewed by a board-certified pathologist the unstained sections are sent for DNA extraction as described in[21]. DNA yield from extraction is obtained using Thermo Fisher Qubit Fluorometric Quantification (Waltham, MA). Samples with less than 25 ng of DNA yield are not sent for MSK-IMPACT sequencing and are inadequate for genomic profiling. *EGFR* status was obtained either by another method (e.g., qPCR or fragment analysis) or another biopsy or resection of the same tumor was performed that had adequate tissue. For this study we identified tumor biopsies that were insufficient for MSK-IMPACT but the *EGFR* status was obtained from another source for the same tumor. Cytology samples were excluded as was done in the in-house cohort. A total of 134 cases from 127 patients were obtained. These are samples that are much more challenging to obtain *EGFR* status by molecular means, even in one of the world's most advanced tumor sequencing laboratories. The performance on this cohort illustrates the opportunity to use the computational approach on small biopsy samples.

### MSK Lung Tissue Segmentation Training Cohort

To train a lung tissue segmentation model, we used whole slide images from 83 primary and resected lung adenocarcinomas. 66 images (80%) were drawn randomly for training and the remaining 17 images (20%) were used for validation. All images were partially annotated at pixel-level by an expert pathologist (J.C.) using an in-house slide viewer[42]. Of the 83 slides used to develop the segmentation model, 16 overlapped with the cohorts used for *EGFR* prediction.

### TCGA FFPE Cohort

The lung adenocarcinoma (LUAD) cohort from The Cancer Genome Atlas (TCGA) including the FFPE digital whole slide images were downloaded from GDC portal (https://portal.gdc.cancer.gov). The histology and scanning quality of the TCGA samples varies widely and required manual review of each individual slide to exclude images that lacked adequate detail for automated analyses. Slides with substantial staining and scanning artifacts for which the resolution at 400X magnification (40x microscope objective digital equivalent) was not adequate to identify clear boundaries between tumor cells were removed from the analysis. Of the original 519 slides, 289 passed the manual curation. We hypothesize that when cytologic characteristics cannot be maintained on maximum magnification, the details necessary for identifying mutation-specific morphologic traits are likely to be absent. To evaluate the impact on performance we documented frequently observed artifacts on TCGA slides, including: (i) large deviations in hematoxylin or eosin staining, (ii) cases with high amount of blur on low magnification (20X, 2X microscope objective digital equivalent) over more than 10% of the tissue per slide, and (iii) cases with necrotic tissue of more than 10%.

## Deep Learning

*EGFR* prediction models were trained from non-overlapping tissue tiles of size 224x224 at 20x magnification, following tissue selection on a slide thumbnail and Otsu thresholding as described in Campanella et al.[43]. Experiments were conducted on MSKCC's high performance computing (HPC) cluster utilizing seven NVIDIA DGX-1 compute nodes, each containing eight V100 Volta GPUs and 8TB SSD local storage. Each model was trained in parallel on a single GPU. OpenSlide[44] (version 3.4.1) was used to extract tiles from WSI files on the fly and PyTorch[45] (version 1.0) was used for data loading, modeling, training and inference.

### Lung Tissue Segmentation

A Deep Multi-Magnification Network (DMMN) was used for lung tissue segmentation[25]. DMMN takes a set of patches from multiple magnifications as input to generate a semantic segmentation of the tissue. Specifically, we used a set of patches of size 256x256 at 20x, 10x, and 5x magnifications to generate a segmentation mask of size 256x256 at 20x magnification. We trained a 13-class model, segmenting various tumor and non-tumor subtypes including acinar, lepidic, micropapillary, solid, papillary, spread through air spaces (STAS), histiocytes, lymphocytes, airspaces, fibrosis, necrosis, pen marker and background. To efficiently annotate multiple tissue subtypes on whole slide images, Deep Interactive Learning (DIaL)[24] was used. After training a model with an initial set of annotations, the pathologist (J.C.) corrected any mislabeled regions and we finetuned the model by adding these corrected regions to the training set. During the initial training and finetuning, we optimized the segmentation models with the weighted cross entropy loss using stochastic gradient descent with a learning rate of $10^{-5}$, a momentum of 0.9 and weight decay of 0.0001, with random data augmentation transformations, and selected the final model based on the highest mean intersection-over-union (IoU) on the validation set.

### Supervised Learning

For each of the 20 train/validation splits, model training was replicated three times and the validation AUC of the last epoch was used to select the best model to report performance for each split. All supervised experiments were performed with a ResNet34 model pretrained on ImageNet. Models were trained for 30 epochs with Stochastic Gradient Decent (SGD) with an initial learning rate of 0.05. The learning rate was decayed by multiplying by 0.1 every 10 epochs. Following the empirical class distribution, positive samples were weighted with a value of 0.7 (vs. 0.3 of the negative class). For each epoch, 10% of the tiles from each slide were randomly sampled.

### Weakly Supervised Learning

We used the gated MIL attention model from Ilse et al.[28]. We tried several training strategies which differed in the way that features were extracted from tiles. As a baseline we extracted features using a truncated ResNet50 trained on ImageNet as described in Lu et al.[27]. Since that approach did not yield satisfying results, we the trained in a fully supervised manner as described in the previous section on half the training/validation cohort (n= 1,249). This supervised learning pretraining was performed with a ResNet34 and a truncated ResNet50. The extracted features were then used as input in the gated MIL attention models. The same train/validation splits as before were used here, by first removing the half of the data that was used for pretraining. All the experiments were performed with the Adam optimizer with an initial learning rate of 0.0001 for 50 epochs. As before, samples were weighted based on the empirical class distribution to correct for class imbalance. For every epoch 10% of the slides were randomly sampled.

To compare performance to another state-of-the-art MIL implementation we performed experiments with transMIL[29]. The same features coming from a truncated ResNet50 trained on half the training data, as used for the GMA model, were used here. The model was trained on the remaining half of the training data with a learning rate of 0.0002 and Adam optimizer for 50 epochs. As before, sample weights were adjusted to correct for class imbalance. For every epoch 20% of the slides were randomly sampled.

### Integration of Other Data Modalities

We designed a new model architecture to integrate clinical covariates. We enhance the GMA model by concatenating the output scores for negative and positive classes to a vector of clinical variables and/or to class scores from additional CNNs for other image modalities such as PET images. This new vector is the input to an additional classification layer resulting in the final scores (see **Figure 1b**). The model was optimized with the same parameterization as the histology-based GMA model.

### Signed Attention

The attention weights from the trained gated MIL attention (GMA) model do not offer information on whether the importance of an instance is towards a negative or positive classification. In order to retrieve this type of information we define the instance sign as $h \cdot w^T > 0$, where $h$ is the feature vector for the particular instance, and $w$ is the weight vector of the classification layer.

## PET-based Model Development and Experimentation

For a subset of 724 cases of the study cohort pre-treatment (within 4 months of treatment start) PET/CT imaging were available. Patients were injected with on average 440 MBq FDG and scanned for 3 min per bed position approximately 1 hour after injection. The acquired PET data was reconstructed according to clinical standard, and the Bq/ml uptake images were all converted to Standardized Uptake Values (SUVs) and resampled to a uniform voxel size of 3.27x3.27x3.27 mm. All PET experiments followed the same dataset partitioning as the histology experiments with 546 of the 724 cases used for training/tuning and 178 for testing. The data was split randomly 20 times into training and validation datasets using the same splits as in the main cohort, and the model was trained separately on each split using class balancing. Three types of models were explored: 1) PET only model using unannotated MIPs, 2) PET only model using 3D tumor boxes, and 3) multimodal model using both histology images and PET MIPs. For the MIP models, one 2D coronal and one 2D sagittal MIP was generated from the 3D volume image, and the images were automatically cropped to approximately mid abdomen to the base of the skull. Each MIP was cropped further to its non-zero bounding box and placed randomly on an empty 220x220 canvas. The entire dataset was clamped at SUV 40 and z-normalized, and the two MIP views were used as a two-channel input to a deep convolutional neural network model. We used a pretrained ResNet34 image feature extractor, followed by a fully connected layer taking the 512 ResNet features as input to a binary output. The ResNet34 was pretrained against censored survival times using the PET images of 3,008 lung cancer patients with known overall survival. For the tumor box model, the main tumor was identified, and its central point was annotated on the 3D PET images by a radiologist (A.S.B.). The image boxes were also clamped at SUV 40 and z-normalized. 143 cases were omitted because they were not clearly delineated against the background (either low uptake of lesion or high surrounding uptake by infectious/inflammatory changes). A total of 473 cases were used for these experiments. A one channel 3D image box around the central point was used in a similar model as for the 2D MIPs but using a 3D ResNet34 instead. Different box sizes were tested, with best results obtained for a size of 40x40x40 pixels. For the multimodal model, the histology score of the gated MIL attention model described earlier was concatenated with smoking status and the score of the PET model using unannotated MIPs. The concatenated scores and smoking status served as input to a two-layer Multilayer Perceptron (MLP) with 10 nodes connected to a final binary output. The full model was trained simultaneously, with a weighted loss $L_{tot} = \alpha L_{hist} + (\alpha/2) L_{PET} + \alpha L_{MLP}$ for $\alpha = 0.4$, where the L terms are the binary cross entropy losses of the histology score, PET score, and MLP combined score, respectively. For all three models, the final prediction was the average over an ensemble of the top-10 models over all training data splits.

## Global Lung Cancer Incidence, *EGFR* Mutation and Testing Rates

We obtained the lung cancer prevalence in different countries from their respective registries. We selected the US, China, Brazil and Germany as examples to capture global trends. The case count was 250,000 in the US in 2020[30], 815,000 in China in 2020[31], 30,200 in Brazil in 2020[32] and 56,000 in Germany in 2016[33]. Data regarding incidence of lung adenocarcinoma is more difficult to obtain since many national registries do not stratify data by morphology. We used the IARC data[34] to estimate the proportion of lung adenocarcinomas in different regions of the world.

Despite *EGFR* being a well-recognized biomarker, data regarding mutation rates is lacking even in countries with a developed molecular testing infrastructure such as the US. Furthermore, reports that try to compile information vary widely in their estimates, thus resulting in a wide range of possible values. It is commonly accepted that east Asian populations have a higher incidence of *EGFR* mutations[35]. According to Graham et al.[36], the percentage of patients with *EGFR* mutations are 9% in the US, 37% in China, 8% in Brazil, and 14% in Germany. In contrast, Midha et al.[37] reports 23% for the US, 48% for China, 28% for Brazil, and 11% for Germany. Given the wide range of reported values, we used both the upper and lower estimates for our analyses.

The rate of molecular testing for *EGFR* mutations is even less known. In the US, testing has been steadily increasing over the years[5]. A recent survey[3] determined the testing rate for *EGFR* alterations to be 72%, while another study found a rate of 76%[4]. While these estimates are reflective of regions that are served by elite medical centers, the situation is likely different for more under-served regions of the country. The disparity in access to molecular testing in the US is well reported[38], especially for next generation sequencing[39]. For regions outside of the US we rely on the data from Pennel et al.[5]. The testing rate for China is between 42% and 46%, 38% for Brazil, and 66% for Germany. **Extended Table 1** summarizes the information gathered regarding cancer incidence, *EGFR* mutations occurrence, and molecular testing statistics.

## Clinical Application: *EGFR* Mutation Screening

We estimate the number of patients every year that receive a sub-optimal treatment, based on the adenocarcinoma occurrence, the *EGFR* mutation rates, and the molecular testing data from the literature. For the *EGFR* mutation rates and the testing rates we employ a low-high range to account for the variability in reported data. Given the number of new lung adenocarcinomas per year $N$, the occurrence of *EGFR* mutations $p_{EGFR}$, and the rate of testing $p_{test}$, the number of people that receive a sub-optimal treatment is simply:

$$N * p_{EGFR} * (1 - p_{test})$$

We pair the lowest estimate for $p_{EGFR}$ with the highest estimate for $p_{test}$ to obtain the lower bound on the number of sub-optimal treated patients and vice versa for the upper bound. As an example, in the US, the lower bound for the number of LUAD patients who receive sub-optimal treatment is $102{,}500 * 0.09 * (1 - 0.76) = 2{,}214$, whereas the upper bound is $102{,}500 * 0.23 * (1 - 0.72) = 6{,}601$.

We studied the effect of applying our predictive model for better distributing molecular testing efforts. First, we estimated the number of molecular tests currently performed $N_{test}$, for which there is a lower and upper bound. Then we calculated the operating point that yields a number of predicted positive candidates closest to the number of tests performed in the population. Given the ROC curve $R$ from the in-house test set, we minimize:

$$\min_{se, sp \in R} |N_{test+} - N_{test}|$$

where $N_{test+} = se * N * p_{EGFR} + (1 - sp) * N * (1 - p_{EGFR})$ is the number of positive tests if applying our model with a set sensitivity $se$ and specificity $sp$.

Based on the retrieved sensitivity and specificity, we calculated how many true *EGFR* positive LUAD patients would be missed if molecular testing was performed on all patients predicted positive by our model:

$$N * p_{EGFR} - pr * N_{test+}$$

where $pr = p_{EGFR} * se / (p_{EGFR} * se + (1 - p_{EGFR}) * (1 - sp))$ is the precision of the model.

Similar to before, the proposed lower bound of sub-optimally treated patients by deploying our model is obtained using the lower bound of the ratio of *EGFR* positive patients and the upper bound of the ratio of molecular tests performed. Instead, the upper bound of sub-optimally treated patients is given by the upper bound of *EGFR* positive patients and the lower bound of the ratio of molecular tests. The operating points for lower and upper bounds, as well as the number of mistreated patients are shown in **Extended Table 3**.

## Application to Clinical Trial Enrollment

We ran simulations to estimate the number of *EGFR* positive patients that can be expected with 95% confidence when sampling from the population of lung adenocarcinoma patients at random versus our predictive model. In each scenario 10,000 trials were run to estimate the confidence bounds. For patient selection utilizing our predictive model, we used a probability threshold of 0.33 which was obtained optimizing the Youden's index[46] on the validation set. Since these estimates are dependent on the occurrence of the *EGFR* mutation in the population, we use the bound analysis as before. Patient enrichment is defined as the ratio of *EGFR* positive patients sampled by our method and the number of *EGFR* positive patients that would be sampled at random.

# Extended Results

## H&E-based Biomarker Development

### Baseline Supervised Learning Experiments

The baseline approach to supervised learning of mutation status in lung cancer is tile level training by assigning the mutation status from the slide level to all tiles of a slide (c.f. Coudray et al.[10], Kather et al.[13]). This approach did not yield satisfactory results on our clinical dataset, with an average validation AUC of 0.73 (**Extended Figure 2a**). To improve over this result, we trained a semantic segmentation model (see **Extended Methods**) to distinguish between non tumor regions and tumor regions. Based on the segmentation we trained the *EGFR* predictor only on tumor tiles resulting in an improved performance with a validation AUC of 0.77 (**Extended Figure 2a**).

The segmentation model was also trained to distinguish between lung adenocarcinoma morphologic subtypes including acinar, papillary, micropapillary, solid, and lepidic (see **Extended Methods**). Having semantic segmentation data on tile level allows for better understanding of model performance in the context of tumor morphologies that are familiar to thoracic pathologists. The tumor specific model showed better performance on predominantly acinar, papillary, and micropapillary tumors while predominant lepidic and solid tumors resulted in worse overall performance (**Extended Figure 2b**).

### Subtype-specific Supervised Learning

To better understand the dependencies of prediction performance on morphological subtypes of lung adenocarcinoma, we independently trained models for each subtype by ignoring all tiles classified as a different subtype by the semantic segmentation model. As seen in the prior experiment, acinar, papillary, and micropapillary morphologies showed superior performance (0.78, 0.75, and 0.76 average validation AUC respectively, **Extended Figure 2a**). Interestingly, the performance of each of these models was improved or on-par with the model that was trained on all tumor tiles, despite utilization of much fewer tumor tiles. In contrast, models trained and tested on solid and lepidic morphologies showed performance that was markedly degraded (0.72 and 0.66 average validation AUC respectively) compared to the model trained on all tiles. These experiments collectively lead to the conclusion that the different morphologic subtypes contain varying levels of discriminatory characteristics, specifically that acinar, papillary and micropapillary subtypes express more differentiating morphologies regarding the *EGFR* mutation status.

The final model is an ensemble of the 10 best performing models on the validation set. Its performance was assessed on each held-out test set by averaging the predictions of the 10 models of the ensemble. The performance on the MSK in-house test set was very similar to the performance observed on the validation sets: AUCs of 0.76, 0.77, 0.77, 0.72, and 0.68 for acinar, papillary, micropapillary, solid, and lepidic respectively (**Extended Figure 2c**). The performance on the MSK external test set was on average 3% lower in terms of AUC compared to the in-house test set: AUCs of 0.73, 0.73, 0.73, 0.73, and 0.60 for acinar, papillary, micropapillary, solid, and lepidic respectively.

In addition, we analyzed the performance in terms of primary and metastatic samples and observed a clear difference between the two. Predictions for primary samples had on average 0.06 higher AUC than predictions for metastatic ones (**Extended Figure 2d**). This held true when considering in-house and external cohorts separately (**Extended Figure 2e-f**). Given that the external test cohort is enriched in metastatic samples, these results explain the discrepancy in performance between in-house and external cohorts.

We explored the possibility of training separate models for primary and metastatic samples. When comparing the performance of models trained jointly (on primary and metastatic samples) with the models trained separately, the former outperformed the latter (**Extended Figure 2g-h**). This could be due in part to the small number of training examples once stratified by primary and metastatic samples.

In some cases, slides may contain only a small number of tiles for a specific sub-type. This could be due to a truly low abundance of that subtype or based on spurious misclassification by the segmentation model. In these cases, making a prediction based on a few tiles may be misguided. By filtering out slides by their number of tiles for each specific subtype we show further performance gains (**Extended Figure 2i-j**).

## Subtype Agnostic Weakly Supervised Learning

To further investigate our hypothesis of morphological sub-type dependence, we implemented the gated MIL attention (GMA) model from Ilse et al.[28], a weakly supervised learning technique which can automatically determine the regions on a slide that contribute the most to a classification. It is important to note that in these experiments, we do not use any knowledge of the subtypes to train the models. As baseline, we utilized the procedure used in Lu et al.[27] by extracting features of tumor tiles from the penultimate block of a ResNet50 (or truncated ResNet50) trained on ImageNet and trained a gated MIL Attention network. The resulting performance was quite poor compared to the fully supervised experiments (**Extended Figure 3a**), highlighting how ImageNet features may not be suitable for certain tasks. To obtain features more relevant to our task, we took half of the training/validation data split (n=1,249) to pretrain a ResNet34 for the task of *EGFR* prediction from all tumor tiles. We then extracted the features learned by this pretrained network for all the tumor tiles and trained a GMA network on the rest of the train/validation data following the same 20 random train and validation splits as described before. The resulting validation performance was on par with that of the supervised experiments based on the acinar subtype (**Extended Figure 3a**). To assess the necessity of a segmentation model for tile selection, we used the same features as before to train a network from all tiles in a slide, instead of just tumor tiles, achieving comparable performance. This approach is preferable because it eliminates the need to rely on a segmentation model during training and inference. To measure improvement with a higher capacity CNN model we repeated the experiments with a truncated ResNet50. Performance was improved for both for training on tumor tiles as well as on training on all tiles. For the rest of the analyses, we used the gated MIL attention model with lung-pretrained truncated ResNet50 features from all tiles of a slide.

As described before, an ensemble of the top-10 performing models was used to analyze the performance on the MSKCC In-house and external test cohorts, resulting in an AUC of 0.79 and 0.77 respectively (**Extended Figure 3b**). These results are well in line with the validation results, demonstrating good generalization performance on both in-house and external cohorts.

To investigate further the hypothesis that certain subtypes show a phenotypic change when an *EGFR* mutation is present and hence are more important for *EGFR* prediction, we analyzed the attention maps generated by the ensemble model on the in-house test cohort. Importantly, we distinguish between attention for a positive and negative *EGFR* prediction (see **Extended Methods**). For this analysis, we consider the median attention value within each subtype for each slide. In **Extended Figure 3i** we observe a clear differential in positive attention for acinar, papillary, and micropapillary subtypes, for *EGFR* mutated slides versus *EGFR* wildtype slides, demonstrating again the importance of these subtypes for predicting *EGFR* status. The negative attention does not show any significant trends except that it is lower for mutated *EGFR* than its wildtype counterpart as expected (**Extended Figure 3j**). It is noteworthy that the lepidic subtype shows lower attention values overall.

Analyzing the model's performance regarding metastatic and primary samples showed a marked drop in AUC from 0.80 to 0.72 for metastatic tumors independent of being in-house or external samples (**Extended Figure 3c**). We hypothesize that this could be due to the more varied morphologies that are encountered in metastatic tumors that may not be well captured in our training dataset. Given that primary tumors grow within the constraints and stromal architecture of the lung, metastatic lesions develop in a broad set of stromal environments from constrained dense stromal architectures of the liver and lymph nodes to more loose stromal environments like the brain. Trying to understand why metastatic samples perform relatively poorly, we stratified the predictions based on metastasis location (**Extended Figure 3e**) but could not recognize any reproducible pattern. When stratifying predictions by smoking status, we observed a definite trend where the *EGFR* status could be detected more easily for current smokers than patients who never smoked (**Extended Figure 3f**). The analysis by cancer stage was hampered by the fact that it was missing for 733 cases out of 1,339. Taking this into account, we did not observe a major difference in performance between early stage (stages 1 and 2) and late stage (stages 3 and 4), independent of being a primary or metastatic sample (**Extended Figure 3g**). Stratification by most prevalent subtype in each slide showed that the most challenging slides for *EGFR* prediction are the ones that are mostly of solid morphology (**Extended Figure 3h**). To determine whether our predictive power comes from a specific type of *EGFR* mutation, we also analyzed the *EGFR* variants and categorized them into the following subsets: Exon 19 deletion, Exon 20 insertion, L858R, T790M, and other *EGFR* mutations. Stratifying by variant, we observed a differential in prediction performance between primary and metastatic samples. Exon 19 deletion, L858R, and T790M were predicted with higher accuracy in primary samples. Overall, prediction of exon 19 deletions showed the worst performance (**Extended Figure 3d**).

Finally, we analyzed the relative importance of this model's score compared to other clinical variables available. To do so, we trained a logistic regression to predict *EGFR* mutation on the in-house test cohort using our ensemble model score and 14 clinical variables as inputs. The coefficients assigned to each input (**Extended Figure 3k**) clearly show that the features captured by our morphology-based model are vastly superior to other clinical variables and statistically significant, independent of the clinical covariates. As expected, smoking status is the second-most important parameter.

## The pitfalls of training models on TCGA

Previous work based on TCGA datasets for mutational status predictions using fully supervised training on tissue tiles reported good predictive results[10]. Furthermore, curating the training tiles by filtering out "squamous-like" morphology was found to be effective in improving performance. We tried to recapitulate those findings by performing extensive experiments on the TCGA FFPE cohort. In line with Coudray et al.[10], we trained a classifier in a supervised manner using all tiles from the TCGA LUAD FFPE slides (n=519). As previously discussed, this is possible due to an inclusion bias of TCGA resulting in abnormally high tumor purity in contrast to sequential clinical samples. We randomly sampled 30 data splits, where 70% of samples were for training, 15% for validation and 15% for testing, to estimate the performance that can be achieved on the TCGA cohort. The results show that for a dataset of this size the spread of results is very wide, ranging from 0.4 to 0.9 AUC on validation and test sets (**Extended Figure 5a**) with no correlation between validation scores and test scores. In addition, replicated experiments on the same data splits achieved near identical results, emphasizing the high degree of dependence on a specific data split.

To reproduce filtering out of "squamous-like" tiles, we trained a LUSC vs. LUAD classifier, that performed with an AUC of 94%. Removing the most "squamous-like" tiles and rerunning the experiments as before on the same data splits, did not lead to an improvement over the previous result (**Extended Figure 5b**). Finally, we restricted the data to the curated set of slides (n=289) and repeated the experiments. Similarly, both validation and test performances varied widely across the 30 random data splits.

In conclusion, the TCGA LUAD cohort is small, and the performance obtained can vary greatly depending on the data split used. Consequently, models trained on TCGA data will suffer from a wide generalization gap when tested in a real-life clinical setting. Furthermore, we could not replicate the finding that squamous filtering can help improve performance. To address these issues, we would like to encourage the community to report replication experiments on multiple random dataset splits when working with slides from the TCGA.

## Multimodal integration of pathology, clinical variables, and other image modalities

It is well known that smoking status can be a predictor for *EGFR* status[47]. We designed a multimodal experiment by expanding our histology-based model with the patients' smoking status and achieved a validation AUC of 0.84 (**Extended Figure 4a**), greatly improving the performance over using either pathology (0.79) or smoking status (0.70) alone (**Extended Figure 1a**). To test the impact of the remaining clinical variables we trained additional models by progressively including them into the model as depicted in **Figure 1b**. The results plotted in **Extended Figure 4a** showed no difference based on the sample coming from a primary or metastatic tumor. Including sex and race together with histology and smoking status improved the average AUC slightly. Additional inclusion of metastasis status, stage, and age at diagnosis did not result in improvements. Even though the model including histology, smoking status, sex and race achieved the best average validation AUC, we propose to follow Occam's razor and use the simpler model including only histology and smoking status, due to the fact that the performance difference is not significant (Mann-Whitney U Test p-value: 0.090).

The performance of the final model including histology and smoking status generalized well to the in-house and external test set as shown in **Extended Figure 4b**. Stratifying the patients by sample type, primary samples perform well in both in-house and external cohort mirroring the results from all previous experiments. The introduction of smoking status does improve the performance on in-house metastatic samples, which is on par with the performance on primary samples (**Extended Figure 4c**). When stratifying by metastasis site we found that liver samples showed reduced performance (**Extended Figure 4d**). Stratifying by stage did not yield any interesting observations (**Extended Figure 4e**). When stratifying by prevalent subtype, we saw the same behavior as in previous experiments where predominantly solid slides yield a lower performance (**Extended Figure 4f**). Finally, when stratifying by *EGFR* mutation variant we observed a similar behavior as for the histology-only model, with exception of an improved performance on the T790M mutation for metastatic sample (**Extended Figure 4g**).

We also investigated the predictive power of FDG PET images for determining the *EGFR* mutation status. We identified 724 cases (n=546 training/tuning, n=178 test) from the study cohort that also had a pre-treatment FDG PET/CT scan and trained a classification model on unannotated 2D PET maximum intensity projection (MIP) images of the torso. Using the same held out test subjects, the ensemble PET model alone reached an AUC of 0.62 (**Extended figure 4h**). We further tested a PET model using smaller 3D boxes centered around the main lung tumor, but the performance was lower than that of the MIP model. In contrast to Mu et al.[18] and Yin et al.[19] who used small tumor-centered regions as model input, we did not achieve an AUC comparable to histology-based models. Finally, we tested a multimodal model combining the histology GMA model and the PET MIP model, as well as smoking status. However, the model performed worse than the model using only histology and smoking status (validation AUC of 0.82 versus 0.84).

# Extended Figures

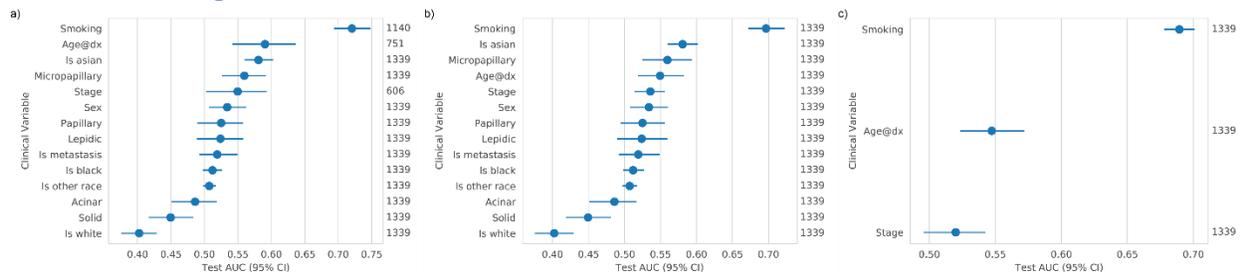

**Extended Figure 1 Performance of EGFR prediction from clinical variables only.** *a) 95% bootstrapped AUC for each clinical variable. Missing values are removed. b) 95% bootstrapped AUC for each clinical variable. Missing values are imputed with the mode. c) Distribution-based random imputation for the three variables with missing values.*

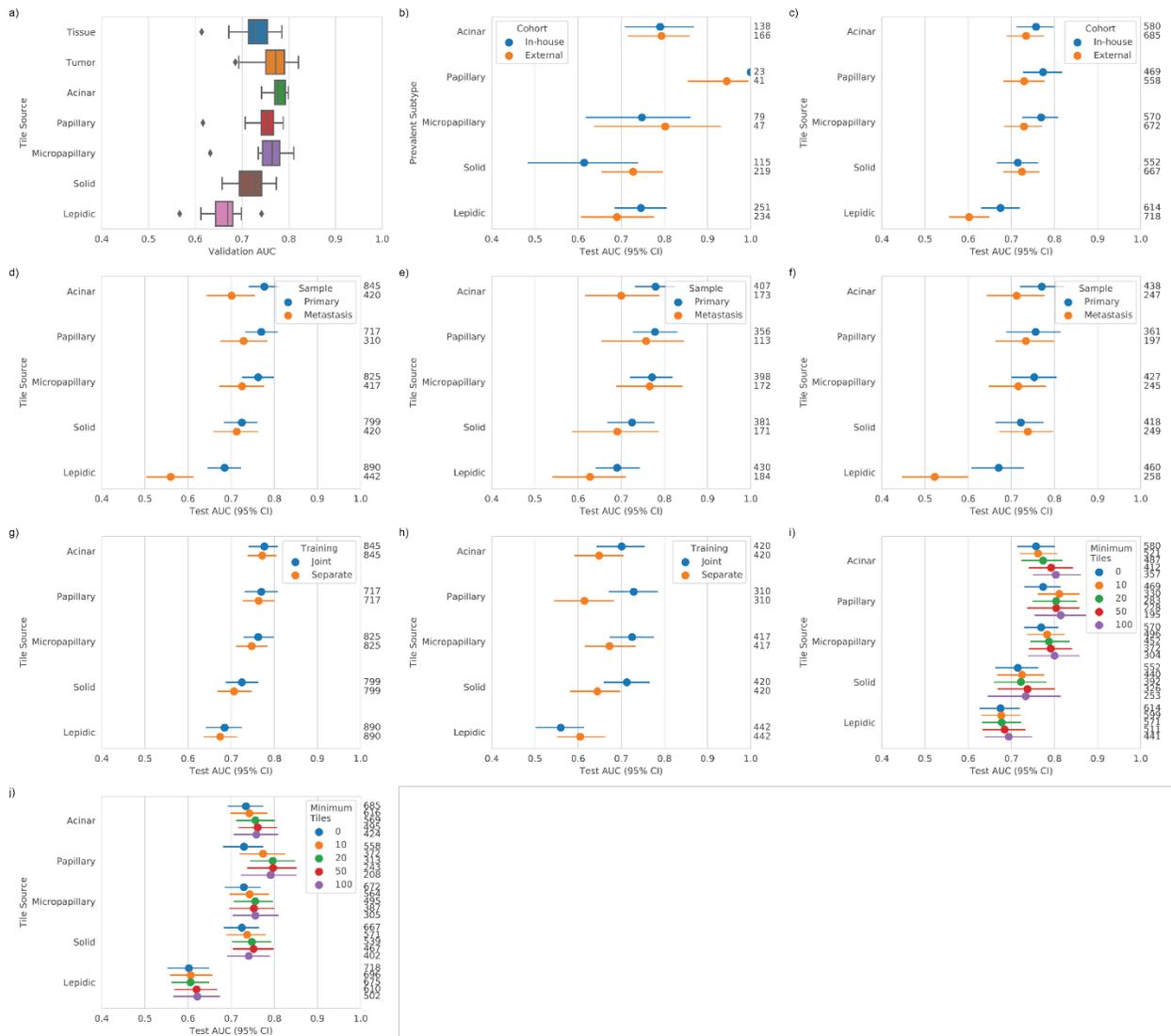

**Extended Figure 2 Results of supervised learning experiments.** *a) Validation performance in terms of AUC distribution from 20 random dataset splits for experiments performed from different subsections of tiles based on morphology. b) Test performance of the supervised model trained on tumor tiles only, stratified by prevalent morphology subtype present in a slide. c) AUCs of top-10 ensemble models trained for each morphological subtype and stratified by test cohort (in-house and external). d) Top-10*

*ensemble test AUCs of subtype specific trained models stratified by sample type (primary or metastatic). e) Top-10 ensemble test AUC (in-house only) of subtype specific trained models stratified by sample type (primary or metastatic). f) Top-10 ensemble test AUCs (external only) of subtype specific trained models stratified by sample type (primary or metastatic). g) Comparison of models trained jointly on primary and metastatic samples versus trained separately only on primary samples. Plotted are the top-10 ensemble test AUCs on primary samples. h) Comparison of models trained jointly on primary and metastatic samples versus trained separately only on metastatic samples. Plotted are the top-10 ensemble test AUCs on metastatic samples. i) Top-10 ensemble test AUCs (in-house only) of subtype specific trained models by filtering out slides that contain less than the specified number of tiles for each subtype. j) Top-10 ensemble test AUCs (external only) of subtype specific trained models by filtering out slides that contain less than the specified number of tiles for each subtype.*

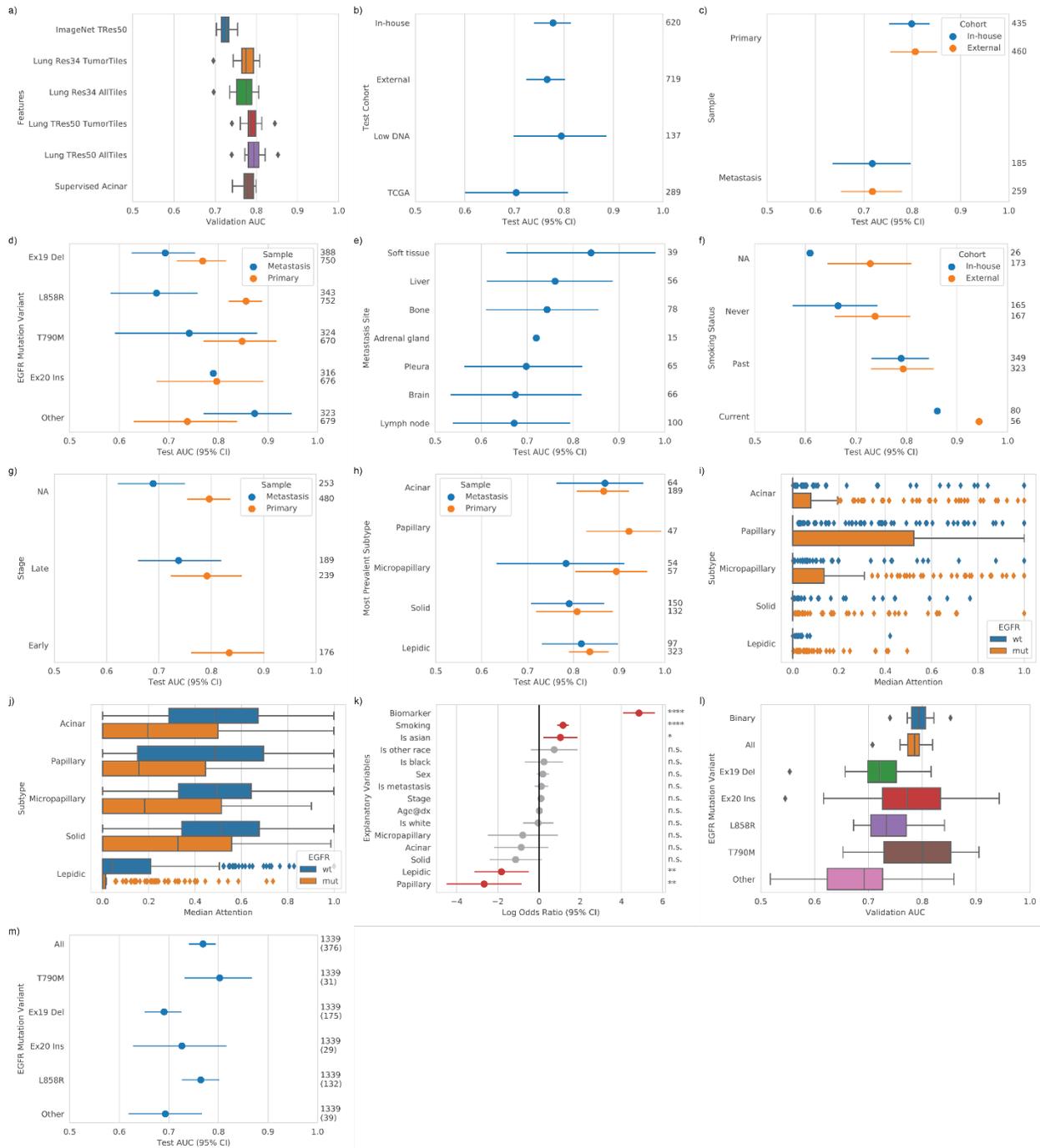

**Extended Figure 3 Results of GMA-based weakly supervised experiments.** a) Validation performance in terms of AUC distribution from 20 random dataset splits for GMA models trained with features extracted with different strategies. The following analyses use the best performing model which was trained with a truncated ResNet50 using all tissue tiles. b) Top-10 ensemble performance on the various test sets. c) Top-10 ensemble performance on the in-house and external sets stratified by primary and metastatic samples. d) Top-10 ensemble performance on the in-house and external sets stratified by EGFR mutation variant and sample type. e) Top-10 ensemble performance on the metastatic samples of the in-house and external sets stratified by metastasis site. f) Top-10 ensemble performance on the in-house and external sets stratified by smoking status. g) Top-10 ensemble performance on the in-house and external sets stratified by tumor stage and sample type. h) Top-10 ensemble performance on the in-house and external sets stratified by most prevalent tumor subtype in each slide and sample type. i-j) Analysis of attention scores (see Extended Methods) produced by the GMA model on the test cohorts. i) Positive attention

*analysis. The median positive attention scores restricted to each tumor subtype were analyzed for wild type and mutated EGFR slides. j) Negative attention analysis. Median negative attention scores restricted to each tumor subtype were analyzed for wild type and mutated EGFR slides. k) Analysis of relative importance of the histology-based GMA model compared to other clinical variables. The output of our model (Biomarker) was used alongside all clinical variables to train a logistic regression for EGFR prediction. The forest plot shows the log odds ratio coefficients and their 95% confidence interval. Statistically significant variables are plotted in red. Abbreviations: n.s.: non significant (p-value>=0.05), \*: 0.01<=p-value<0.05, \*\*: 0.001<=p-value<0.01, \*\*\*\*: p-value<0.0001. l) Validation performance in terms of AUC distribution from 20 random dataset splits for a multitask model trained to jointly predict all EGFR mutations and specific variants. Comparison with the binary model from previous experiments. m) Performance analysis on variants of EGFR mutation. Top-10 ensemble test performance of the multitask model. Number of positive samples is reported in parenthesis.*

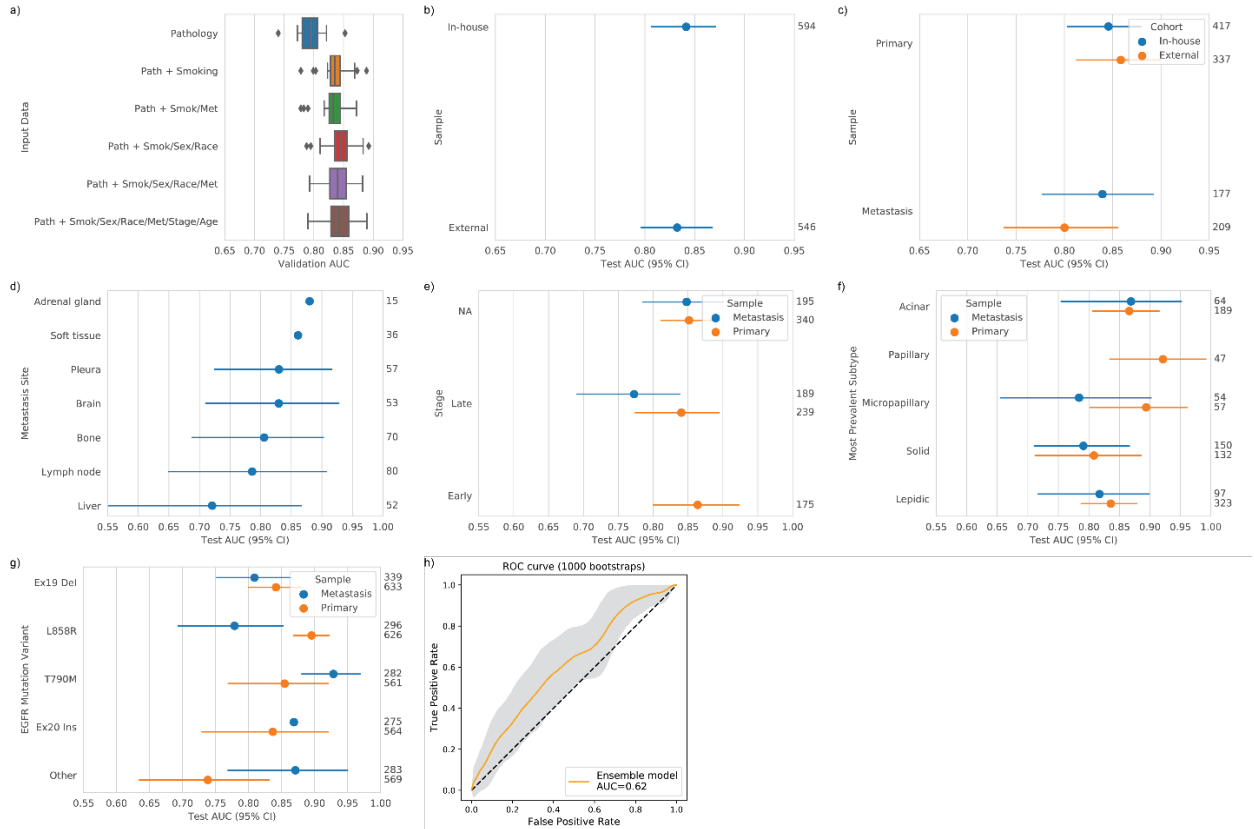

**Extended Figure 4 Results of GMA-based multimodal experiments.** *a) Validation performance in terms of AUC distribution from 20 random dataset splits for GMA multimodal models trained with different subsets of clinical variables. Comparison with the histology-only GMA result. All following analyses use the histology + smoking status model. b) Top-10 ensemble performance on the in-house and external test sets. c) Top-10 ensemble performance on the in-house and external sets stratified by primary and metastatic samples. d) Top-10 ensemble performance on the metastatic samples of the in-house and external sets stratified by metastasis site. e) Top-10 ensemble performance on the in-house and external sets stratified by tumor stage and sample type. f) Top-10 ensemble performance on the in-house and external sets stratified by most prevalent subtype in each slide and sample type. g) Top-10 ensemble performance on the in-house and external sets stratified by EGFR mutation variant and sample type. h) Test performance of a ResNet34 model trained to predict EGFR status from PET images.*

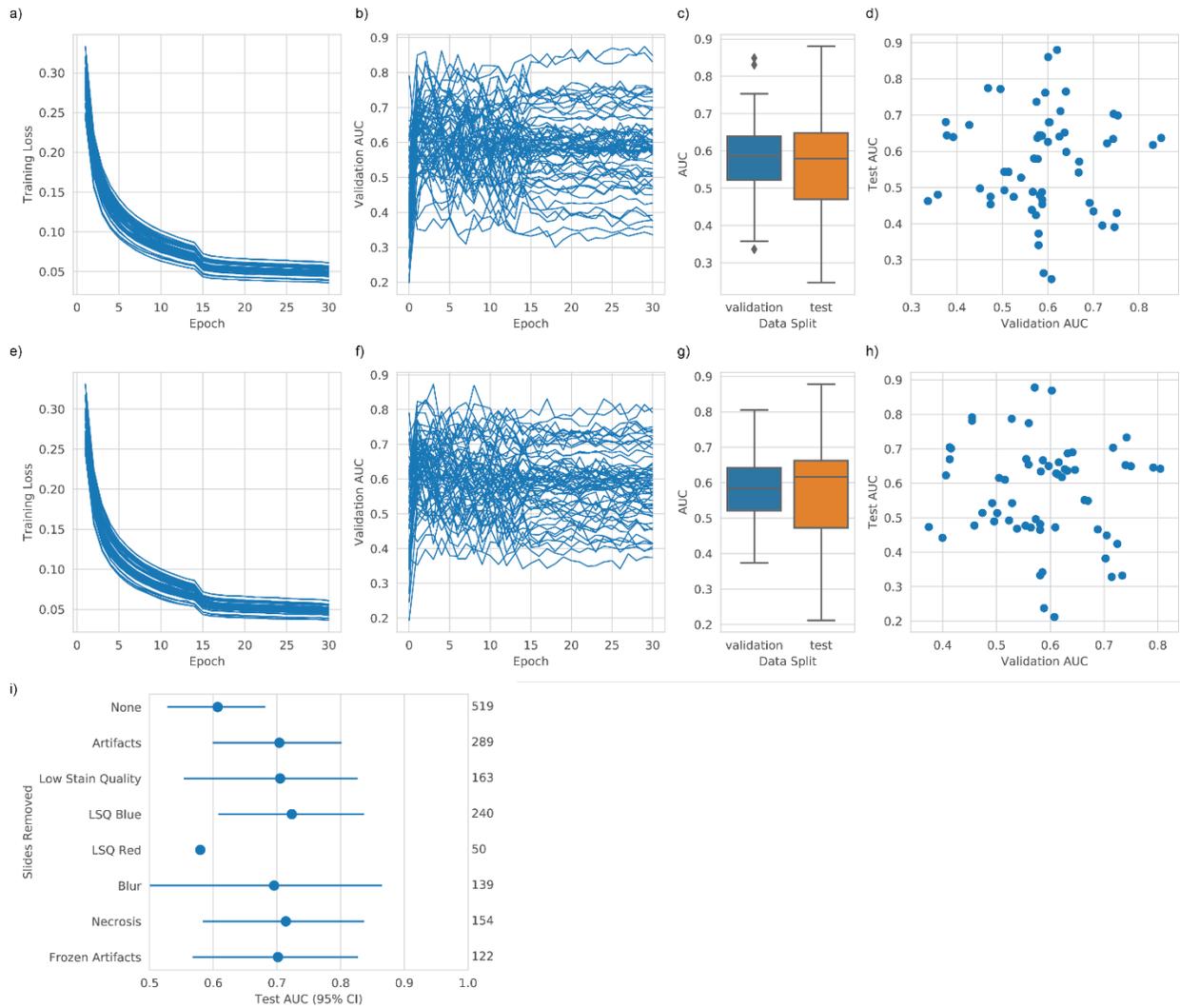

***Extended Figure 5 Results of TCGA experiments.*** *a-d) Supervised experiments on all tissue tiles from all TCGA slides (n=519). a) Training curves of all 30 randomized data sets and their two replicas. b) Validation AUC over training epochs for all runs. c) Boxplots of validation and test AUCs for all runs. d) Scatter plot of validation AUC paired to its respective test AUC. e-h) Supervised experiments restricted to tiles predicted as adenocarcinoma by a squamous carcinoma vs. adenocarcinoma binary classifier. e) Training curves of all 30 splits and their two replicas. f) Validation AUC over training epochs for all runs. g) Boxplots of validation and test AUCs for all runs. h) Scatter plot of validation AUC paired to its respective test AUC. i) Ablation analysis of the performance of the GMA model trained on MSK data. Test AUC for different subsets of the TCGA cohort based on the curation performed by an expert pathologist (C.V.). For the models below "Artifacts", each row represents a model with slides of that condition removed in addition to the slides removed in "Artifacts".*

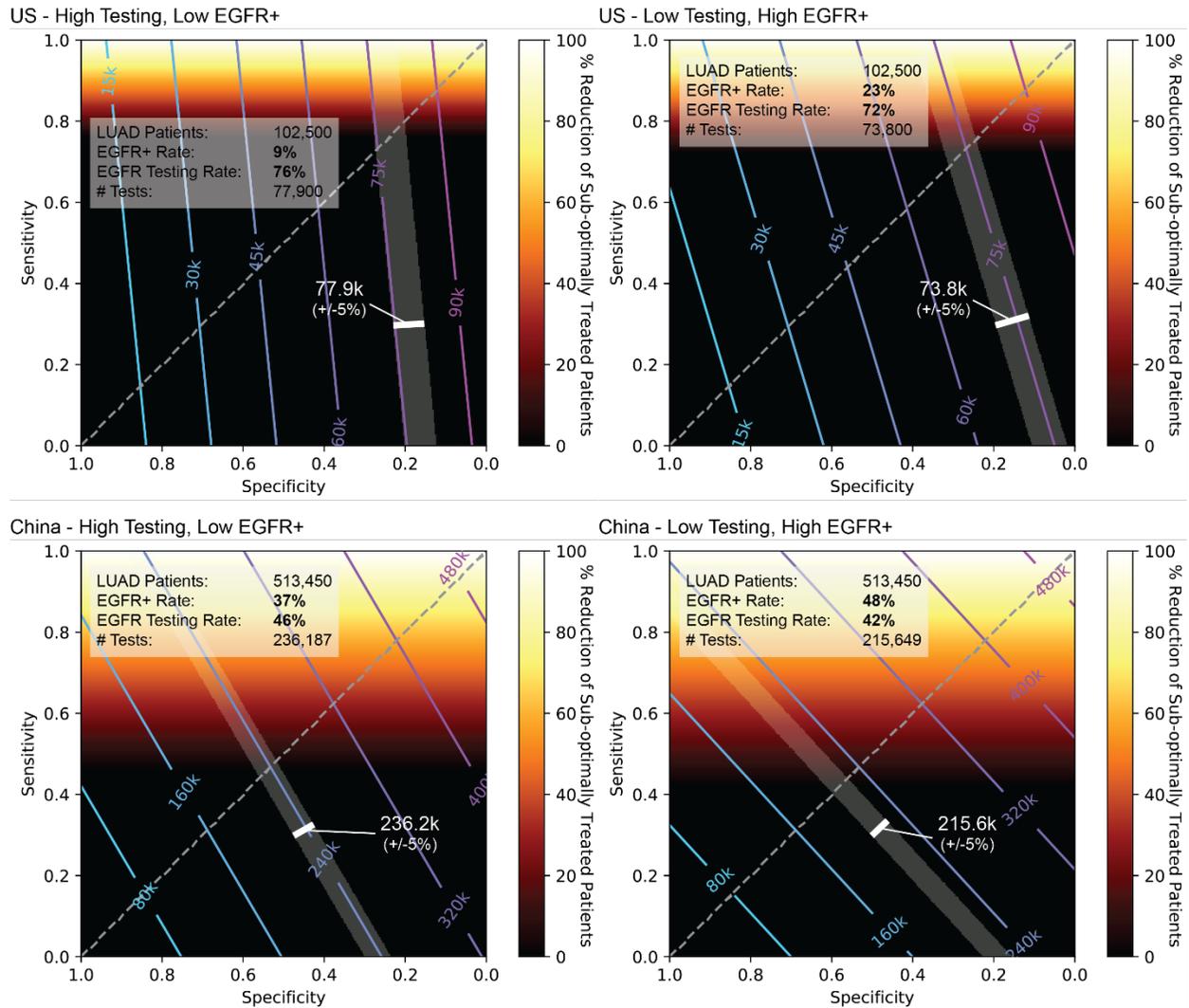

*Extended Figure 6 Clinical application of computational screening for EGFR mutations. Analysis of the hypothetical reduction of sub-optimal treatments by applying an EGFR screening tool with a certain sensitivity and specificity. For every panel, the heatmap represents the percentage reduction of sub-optimal treatments, with the isolines showing the number of positive screening tests. The reduction of sub-optimal treatments depends on the sensitivity of the model, while the number of positive tests depends on both sensitivity and specificity in a manner proportional to the EGFR mutation rates. A region of the surface that represents the current levels of molecular testing is highlighted. Insets provide information regarding each region in terms of number of LUAD patients, EGFR mutation and testing rates. For each US and China, two conditions are presented based on low and high bounds for the estimates of EGFR mutations and EGFR molecular testing. Top left: US with high estimate for testing and low estimate for mutation prevalence. Top right: US with low estimate for testing and high estimate for mutation prevalence. Bottom left: China with high estimate for testing and low estimate for mutation prevalence. Bottom right: China with low estimate for testing and high estimate for mutation prevalence.*

# Extended Tables

## Extended Table 1 Summary of cohorts included in this study

| Cohort | Split | Slides | Primary | Metastatic | Patients | *EGFR*+ (%) |
|---|---|---|---|---|---|---|
| **MSK Segmentation** | Train/Validation | 83 | - | - | 83 | - |
| **MSK In-house** | All | 3,069 | 2,201 | 868 | 2,573 | 31.4 |
| **MSK In-house** | Train/Validation | 2,449 | 1,766 | 683 | 2,056 | 31.7 |
| **MSK In-house** | Test | 620 | 435 | 185 | 517 | 30.3 |
| **MSK External** | Test | 719 | 460 | 259 | 656 | 26.1 |
| **MSK Insufficient DNA** | Test | 134 | - | - | 127 | 20.9 |
| **TCGA FFPE** | Test* | 519 | - | - | 519 | |
| **Curated TCGA FFPE** | Test* | 289 | - | - | 289 | |
| **Total** | | 4,508 | | | 3,875 | |

## Extended Table 2 Lung cancer statistics in the literature

| Country | Lung Cancers | LUAD (%)[34] | LUAD Cancers | *EGFR*+ (%) | *EGFR* Test (%) |
|---|---|---|---|---|---|
| **US** | 250,000[30] | 41 | 102,500 | 9[36]-23[37] | 72[3]-76[4] |
| **China** | 815,000[31] | 63 | 513,450 | 37[36]-48[37] | 42-46[5] |
| **Brazil** | 30,200[32] | 38 | 11,476 | 8[36]-28[37] | 38[5] |
| **Germany** | 56,000[33] | 40 | 22,400 | 11[37]-14[36] | 66[5] |

## Extended Table 3 Sub-Optimal Treatments (SOT) before and after deployment of a universal *EGFR* screening tool

**Sub-Optimal Treatment: Low Bound (Low *EGFR* – High Testing)**

| Country | Sensitivity | Specificity | SOT Before | SOT After | % Reduction |
|---|---|---|---|---|---|
| **US** | 0.992 | 0.263 | 2,214 | 76 | 96.6 |
| **China** | 0.852 | 0.770 | 102,587 | 28,029 | 72.7 |
| **Brazil** | 0.902 | 0.665 | 569 | 90 | 84.1 |
| **Germany** | 0.967 | 0.378 | 838 | 81 | 90.4 |

**Sub-Optimal Treatment: High Bound (High *EGFR* – Low Testing)**

| Country | Sensitivity | Specificity | SOT Before | SOT After | % Reduction |
|---|---|---|---|---|---|
| **US** | 0.974 | 0.356 | 6,601 | 612 | 90.7 |
| **China** | 0.728 | 0.864 | 142,944 | 67,036 | 53.1 |
| **Brazil** | 0.836 | 0.797 | 1,992 | 527 | 73.6 |
| **Germany** | 0.967 | 0.390 | 1,066 | 103 | 90.4 |